\documentclass[10pt,twocolumn,letterpaper]{article}

\usepackage{cvpr}              

\usepackage{microtype}

\usepackage{amsmath, amssymb}
\usepackage{graphicx}
\usepackage{booktabs}
\usepackage{multirow}
\usepackage{amsthm}
\usepackage{makecell}

\usepackage[table]{xcolor}
\usepackage{arydshln}
\usepackage{tabularx}
\usepackage{amsmath} 
\usepackage{colortbl}
\usepackage{siunitx}

\usepackage{algorithm}
\usepackage{algorithmic}

\renewcommand{\paragraph}[1]{\vspace{.5em}\noindent\textbf{#1.}}

\usepackage{xspace}
\newcommand{\FRAP}{Fused Reference Alignment Prediction\xspace}
\newcommand{\SAS}{Semantic Alignment Score\xspace}

\definecolor{R1Color}{RGB}{0, 140, 0}  
\definecolor{R2Color}{RGB}{200, 50, 150}
\definecolor{R3Color}{RGB}{139, 0, 139}  

\definecolor{concern}{RGB}{80, 100, 200}
\definecolor{flag}{RGB}{0, 0, 205}

\definecolor{cvprblue}{rgb}{0.21,0.49,0.74}
\usepackage[pagebackref,breaklinks,colorlinks,allcolors=cvprblue]{hyperref}

\def\paperID{} 
\def\confName{CVPR}
\def\confYear{2026}

\title{\vspace{-0.5cm}Bridging Domain Expertise and Generalization for Performance Estimation}

\author{
Shuxuan Li$^{1}$ \quad
Zhilin Zhao$^{1,2,3}$\thanks{Corresponding author. Code is available at \url{https://github.com/NuyoahNasuS/FRAP}} \quad
Quyu Kong$^{4}$ \quad
Wei-Shi Zheng$^{1,2,3}$ \\
\fontsize{6.5pt}{7.5pt} $^{1}$School of Computer Science and Engineering, Sun Yat-sen University, China \\
\fontsize{6.5pt}{7.5pt} $^{2}$Key Laboratory of Machine Intelligence and Advanced Computing, Ministry of Education, China \\
\fontsize{6.5pt}{7.5pt} $^{3}$Shenzhen Loop Area Institute, China \quad 
$^{4}$Alibaba Cloud
\\
{\tt\small lishx87@mail2.sysu.edu.cn, zhaozhlin@mail.sysu.edu.cn} \\ [-2pt]
{\tt\small quyu.kong@alibaba-inc.com, wszheng@ieee.org}
}

\begin{document}
\maketitle
\begin{abstract}
    Performance estimation under distribution shift aims to predict how a model behaves on an unlabeled test set whose distribution differs from the training data, a scenario that requires reliable indicators that can faithfully reflect model behavior without ground-truth labels. Existing approaches rely solely on the outputs of the given model whose biases are amplified once the distribution shifts, weakening the correlation with the true performance. Motivated by this limitation, we propose \textbf{F}used \textbf{R}eference \textbf{A}lignment \textbf{P}rediction (FRAP), which leverages the complementary strengths of an external foundation model and the base model to construct a more reliable surrogate of the ground-truth labels. FRAP aligns the prediction distribution of the foundation model with that of the base model by applying temperature-scaled calibration that minimizes their divergence. The aligned predictions are fused through confidence-based weighting into a refined reference distribution that integrates robustness from the foundation model and domain-specific expertise from the base model, and performance estimation is obtained by measuring how closely the base model predictions agree with this reference. Extensive experiments across diverse datasets and architectures show that FRAP provides consistent and substantial improvements over representative performance-estimation methods under distribution shift.
\end{abstract}
    
\section{Introduction}
\label{sec:intro}

In the conventional machine learning paradigm, models are trained on a labeled source dataset and evaluated on an unlabeled target dataset. This process implicitly relies on the \iid assumption, which requires both datasets to originate from the same underlying distribution. However, real-world deployment frequently violates this assumption, resulting in significant performance degradation~\cite{Koh2020WILDSAB,Recht2019DoIC}. Consequently, estimating model performance on unlabeled data under distribution shift is essential for building safe and reliable machine learning systems.


In the absence of target labels, performance estimation requires an indicator that reliably reflects the true behavior of the model. Prior work explores a variety of heuristics, including prediction agreement~\cite{Baek2022AgreementontheLinePT,Rosenfeld2023AlmostPE,Yu2022PredictingOE,Jiang2021AssessingGO}, confidence-based statistics~\cite{Garg2022LeveragingUD,Deng2023ConfidenceAD,Guillory2021PredictingWC,Press2024TheEE}, and prediction-distribution characterization~\cite{Lu2023CharacterizingOE,xie2023importance}. Despite methodological differences, these approaches all depend solely on outputs produced by the model itself. Under distribution shift, however, this dependency becomes problematic, \eg, confidence no longer reliably correlates with accuracy, and agreement-based scores may increase even when predictions are consistently incorrect~\cite{Roschewitz2023DistanceMF,Jiang2021AssessingGO,Ovadia2019CanYT,Guo2017OnCO}. These failures stem from limited generalization of the model and the biased predictions that emerge when the input distribution changes. This situation highlights the need for an external source of knowledge that can provide validation beyond the potentially biased outputs of the model under analysis.

A natural solution is to incorporate an external generalized reference that remains independent of the model under analysis and supplies broader knowledge that extends far beyond the source domain. Foundation models naturally satisfy this requirement because training on large-scale diverse datasets enables strong cross-domain generalization. At the same time, the base model retains domain-specific expertise derived from task-oriented training, although its generalization ability diminishes once the input distribution shifts. These complementary properties indicate that integrating a broadly trained foundation model with a task-specialized base model can provide a more stable and informative reference signal for performance estimation. Achieving this integration is nontrivial because the two prediction distributions arise from different training paradigms and feature spaces, and models frequently exhibit notable miscalibration under distribution shift. These challenges motivate a principled mechanism that establishes a consistent probabilistic space for both models and combines their predictions in a reliability-aware manner.

Building on this insight, we propose the \FRAP (FRAP) framework, which unifies strengths from both the foundation model and the base model. FRAP adaptively aligns the prediction distribution generated by the foundation model with the distribution produced by the base model through minimization of Jensen-Shannon (JS) divergence~\cite{Lin1991DivergenceMB}, thereby establishing a consistent probabilistic space and enhancing robustness against potential miscalibration. FRAP subsequently fuses the aligned predictions through confidence-based weighting to obtain refined predictions that incorporate both strong generalization and domain-specific expertise. These fused predictions serve as reference for assessing predictions from the base model through their consistency with the reference. Extensive experiments across diverse datasets and architectures demonstrate that FRAP generally outperforms representative baselines under distribution shift.

In general, our contributions are summarized as follows.
\begin{itemize}
    \item Proposition of FRAP, a novel paradigm for performance estimation on unlabeled datasets under distribution shift.
    \item Design of a pipeline that synergistically integrates the generalization of the foundation model with the domain-specific expertise of the given model.
    \item Deeper investigation and additional experiments are conducted to verify the feasibility of FRAP.
\end{itemize}

\section{Related Works}
\label{sec:Related}

\subsection{Accuracy Estimation}
Estimating model accuracy on unlabeled out-of-distribution dataset, while conceptually simple, remains highly challenging due to the black-box nature of neural networks and the data-dependent behavior of their predictions. These characteristics make rigorous theoretical modeling difficult, leading existing approaches to rely primarily on empirical heuristics extracted from model outputs. Prior work explores signals based on prediction agreement~\cite{Baek2022AgreementontheLinePT,Jiang2021AssessingGO}, confidence statistics~\cite{Rosenfeld2023AlmostPE,Deng2023ConfidenceAD,Guillory2021PredictingWC}, and prediction-distribution shifts~\cite{Yu2022PredictingOE,Lu2023CharacterizingOE,xie2023importance,Press2024TheEE}. Despite methodological diversity, these methods share a key limitation: they derive all evidence exclusively from the model under evaluation. This reliance becomes increasingly unreliable under distribution shift due to biased prediction behavior. The approach introduced in this work departs from this paradigm by incorporating a robust foundation model with strong cross-domain generalization to supply auxiliary guidance, allowing more stable estimation signals under shift.

\subsection{Source Free Domain Adaptation}
Unsupervised Domain Adaptation (UDA) aims to improve the generalization of model to a target domain without target labels, typically through distribution alignment~\cite{Long2015LearningTF,Tzeng2014DeepDC}, adversarial training~\cite{Ganin2014UnsupervisedDA,Tzeng2017AdversarialDD}, or self-training with pseudo labels~\cite{Zou2018UnsupervisedDA,Xie2019SelfTrainingWN}. Source-Free Domain Adaptation (SFDA) extends this setting by removing access to source data during adaptation~\cite{Kim2020DomainAW,Li2020ModelAU,Kundu2020UniversalSD,Liang2020DoWR}, making explicit feature alignment infeasible and increasing reliance on pseudo-labeling in the target domain. Recent advances improve pseudo-label quality through generative assumptions~\cite{Lee2022ConfidenceSF} or by using generalizable multimodal knowledge such as CLIP~\cite{Tang2024ProxyDF}. Although conceptually related through the use of external generalization signals, SFDA focuses on adaptation, whereas the objective in this work is performance estimation on unlabeled data. This fundamental difference leads to distinct design principles and methodological requirements.
\begin{figure*}[t]
    \centering
    \includegraphics[width=\textwidth]{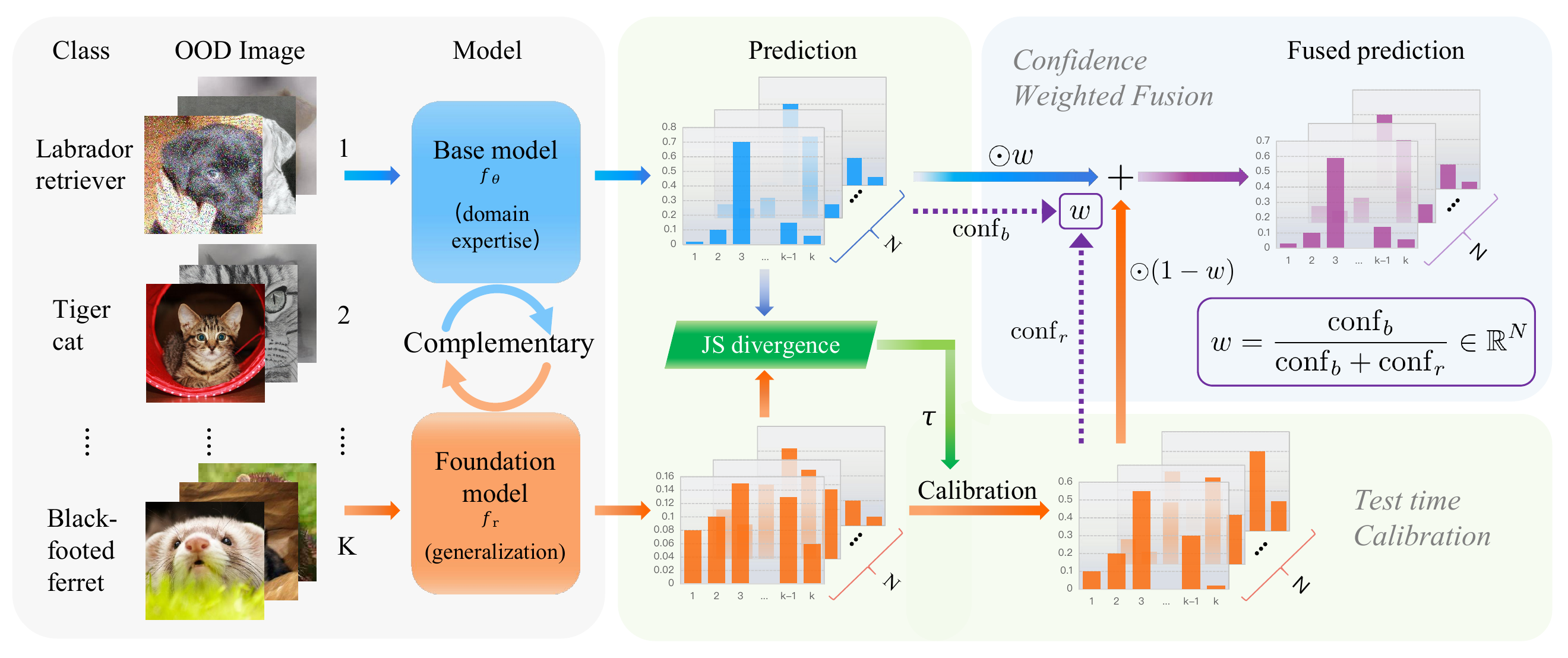}
    \caption{\textbf{FRAP Overview}. FRAP framework leverages a source-trained base model $f_\theta$ together with a robust foundation model $f_r$. The prediction from the foundation model is first calibrated via temperature scaling $\tau$ guided by the output of the base model during test time, and is subsequently fused with the base model prediction through a confidence-weighting scheme. The resulting refined predictive distribution acts as a surrogate for the true label distribution.}
    \label{fig:pipline}
    \vspace{-0.5em}
\end{figure*}

\section{Preliminaries}
\label{sec:Preliminaries}
This section introduces the necessary notation and formalizes the performance estimation problem considered in this work. Building on these definitions, we then present the theoretical basis that motivates the FRAP framework.

\subsection{Problem Formulation}
We focus on the standard image classification setting, where each input-label pair corresponds to an image and its associated semantic category. Let $\mathcal{X}$ denote the input image space and $\mathcal{Y}$ the label space. The source domain follows distribution $\mathcal{D}_s$, while test data are drawn from a different distribution $\mathcal{D}_t$. A classifier $f_\theta$ trained on $\mathcal{D}_s$, is typically evaluated on $\mathcal{D}_t$, often incurring substantial performance degradation~\cite{Koh2020WILDSAB,Recht2019DoIC}. Here, $\theta$ represents the full set of trainable parameters within the neural network classifier $f_\theta$. For clarity, this classifier is referred to as the base model.

Reliable deployment requires accurate estimation of model performance on unlabeled test data. In this study, we address performance estimation under covariate shift~\cite{Heckman1979SampleSB,Shimodaira2000ImprovingPI}, where the label space remains unchanged across domains, i.e., $\mathcal{Y}_s = \mathcal{Y}_t = \{y_1,\dots,y_K\}$ where $K$ indicates the number of classes. Given the labeled source dataset, the unlabeled test dataset, and the trained base model, the central question becomes:
\textit{How can one reliably estimate the performance of this model on the target domain without access to ground-truth labels?}

\subsection{Reformulation of Accuracy}
For any input $x \in \mathcal{X}$, the base model $f_\theta$ outputs a predictive distribution $\widehat{P}_\theta(\cdot \!\mid\! x)$ over the $K$ classes, lying in the $(K\!-\!1)$-simplex. The predicted pseudo-label is defined as $\widehat{y}(x) = \arg\max_j \widehat{P}_\theta(j \mid x)$. For target-domain samples $\{(x_i, y_i)\}_{i=1}^N$ drawn from $\mathcal{D}_t$, the empirical accuracy is
\begin{equation*}
    \mathrm{ACC}(\mathcal{X}_t, f_\theta) \;=\; \frac{1}{N}\sum_{i=1}^N \mathbb{I}\{\widehat{y}(x_i)=y_i\},
\end{equation*} 
where $\mathbb{I}\{\cdot\}$ is the indicator function. 
By introducing the one-hot ground-truth label distribution $P^*(\cdot \!\mid\! x_i) \in \{0,1\}^K$ for sample $x_i$ with $P^*(j\,|\,x_i)=\mathbb{I}\{j=y_i\}$, The expected accuracy can be approximated as following reformulation.
\begin{equation}
\label{eq:root}
    \mathbb{E}[\mathrm{ACC}] \;\approx\; \frac{1}{N}\sum_{i=1}^N \sum_{j=1}^K \widehat{P}_\theta (j \!\mid\! x_i)\,P^*(j \!\mid\! x_i).
\end{equation}

A complete proof is provided in \cref{app:proof}. This formulation provides a natural proxy for performance estimation: the predicted distribution is accessible via forward evaluation of the base model, whereas the ground-truth distribution is unknown in the target domain. FRAP addresses this by constructing a fused distribution that serves as an effective surrogate for $P^*(\cdot\,|\,x)$.

\section{\FRAP}
\label{sec:Methods}

This section presents the construction of the fused distribution that serves as a surrogate for ground-truth labels. The motivation arises from the observation that the base model encodes valuable domain-specific expertise acquired through task-oriented training~\cite{Yosinski2014HowTA,Pan2010ASO}, yet its predictions become biased or unreliable once the input distribution shifts~\cite{Hendrycks2020TheMF}. Foundation models, on the other hand, demonstrate strong generalization across diverse distributions~\cite{Bommasani2021OnTO,Fang2022DataDD,Tu2024ACL}. These complementary characteristics indicate that integrating the broad generalization of the foundation model with the task-specific knowledge of the base model can produce a more faithful proxy for the unknown label distribution.

FRAP puts this idea into practice by combining predictions from the two models through confidence-based weighting, where calibrated confidence provides a natural indication of prediction reliability. Direct fusion, however, is not effective because the two models output confidence values on incompatible scales due to their differing training paradigms. Furthermore, the foundation models used in this work, \ie, CLIP~\cite{Radford2021LearningTV} and SigLIP~\cite{Zhai2023SigmoidLF}) typically produce raw cosine similarities that are poorly calibrated for direct probabilistic use. Without proper scaling, these raw scores fail to reflect true predictive accuracy, a challenge that also applies to other potential foundation models. FRAP addresses these issues by applying test-time calibration to align the prediction distribution generated by the foundation model with the distribution produced by the base model before fusion. The aligned output establishes a shared probabilistic space that enables meaningful confidence-weighted integration. An overview of the FRAP workflow is provided in~\cref{fig:pipline}.

\subsection{Confidence-Weighted Fusion}
\label{sec:module2}
A simple way to combine two model predictions is to take their uniform average. However, this ignores that confidence scores naturally encode how much each model should be trusted. Under proper calibration, higher confidence should correspond to higher predictive certainty and a greater probability of correctness. Motivated by this, we view fusion as selecting, for each input $x$, a single distribution $P^{\ast}(\cdot \!\mid\! x)$ in the label probability simplex $\Delta^K \!=\! \{Q \in \mathbb{R}_+^K \!:\! \sum_{j=1}^K Q(j) = 1\}$ that interpolates between the two model predictions according to their confidence. The confidence associated with the base model and foundation model is defined as
\begin{equation*}
    c_b(x) = \max_j \widehat{P}_\theta(j \!\mid\! x), \quad
    c_r(x) = \max_j P_{\mathrm{r}}(j \!\mid\! x),
\end{equation*}
and is converted into mixture weights over the two models by
\begin{equation*}
    w_m(x) = \frac{c_m(x)}{\sum_{m' \in \{b,r\}} c_{m'}(x)}, \quad m \in \{b,r\}.
\end{equation*}
In particular, the weight of the base model is $\alpha(x) = w_b(x)$ and the weight of foundation model is $1-\alpha(x) = w_r(x)$. For notational convenience, let $P_b(\cdot \!\mid\! x) = \widehat{P}_\theta(\cdot \!\mid\! x)$, the fused distribution is then defined implicitly as
\begin{equation*}
    P^{\ast}(\cdot \!\mid\! x) \in \arg\min_{ Q(\cdot) \in \Delta^K}\!
    \sum_{m \in \{b,r\}}\! w_m(x)\,
    \bigl\lVert Q(\cdot) - P_m(\cdot \!\mid\! x) \bigr\rVert_2^2,
\end{equation*}
and is adopted as a surrogate for the ground-truth label distribution.

The effectiveness of this confidence-weighted fusion critically relies on both models being reasonably calibrated. In practice, this assumption is violated for two fundamental reasons. First, the confidence scales of the two models differ substantially due to their distinct training paradigms. The base model, typically trained with supervised cross-entropy loss on task-specific data, tends to output sharply peaked probability distributions. In contrast, foundation model is trained with a contrastive objective on large-scale image–text pairs from diverse domains, and consequently produces noticeably flatter predictive distributions. This mismatch in confidence scales can induce biased weighting in the fusion, causing one model to be systematically over- or under-weighted regardless of its actual predictive quality.

\begin{figure}[t]
\begin{center}
    \includegraphics[width=\linewidth]{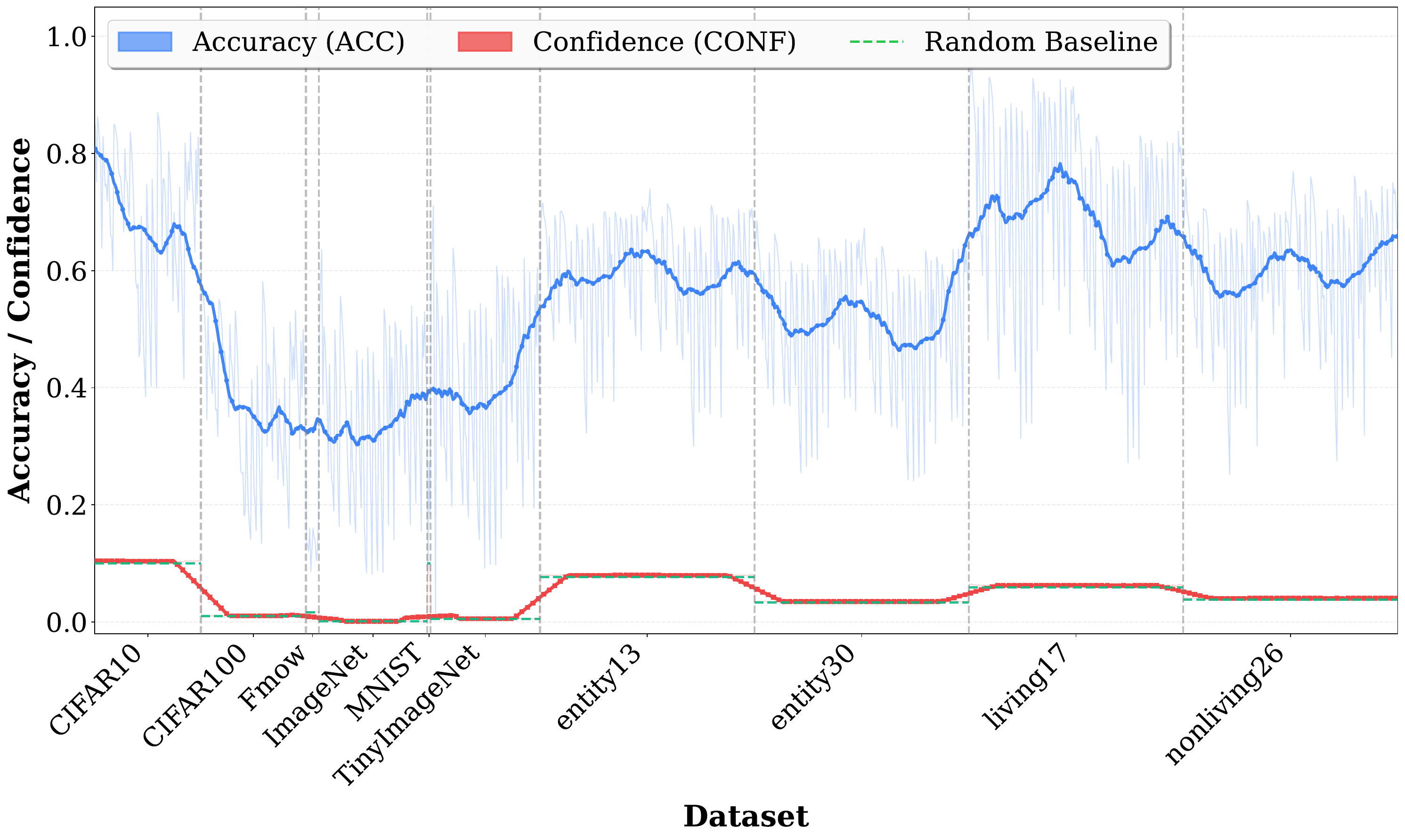}
\end{center}
\vspace{-1em}
\caption{\textbf{Calibration behavior of raw CLIP similarity scores across diverse datasets.} The green dashed line indicates the random baseline corresponding to uniform prediction over the label space (nearly overlapping with the red confidence curve).}
\label{fig:clip_miscalibration}
\vspace{-0.8em}
\end{figure}

Second, the raw outputs from foundation models, such as CLIP, are often poorly calibrated for direct use as probabilistic confidence. 
As illustrated in \cref{fig:clip_miscalibration}, when using the raw cosine similarities between image and text embeddings, the resulting predictive distributions are often close to uniform over the label space. Even for correctly classified examples, the top-1 probability is only slightly higher than the probabilities assigned to other labels. 
This systematic under-confidence severely limits the informativeness as fusion weights and can be attributed to the intrinsic trade-off of contrastive learning objectives \cite{Dinu2025EffectivePE,Schrodi2024TwoEO,Wang2020UnderstandingCR} between uniformity and alignment, which we provide additional details in \cref{app:contrastiveLearning}.

\subsection{Test Time Calibration}
\label{sec:module1}

To address the miscalibration issue identified above, we perform test-time calibration on predictions produced by foundation models (\eg CLIP and SigLIP). A straightforward baseline is to apply a fixed small temperature that rescales the predictive distribution of such foundation models and compensates for suppressed confidence~\cite{Guo2017OnCO}. However, this baseline is inherently data dependent, i.e., an appropriate temperature cannot be chosen without access to labeled target data, and a temperature tuned on labeled source data often fails to transfer under distribution shift.

Therefore, we adopt a dynamic calibration scheme that uses the base model as an adaptive reference for foundation models at test time. Concretely, we align the prediction distribution of foundation models with the prediction distribution of the base model by minimizing their divergence, which yields an unsupervised calibration procedure that automatically adapts to the test dataset. Under distribution shift, the base model typically becomes somewhat over-confident~\cite{Ovadia2019CanYT,Minderer2021RevisitingTC}, yet its prediction distribution still provides a much more informative confidence scale than the nearly uniform probabilities produced by raw similarity of foundation models. The proposed strategy therefore uses the base model as a natural anchor and enforces compatible confidence scales between the two models, which is crucial for effective confidence-weighted fusion.

Formally, a temperature parameter $\tau$ is introduced to rescale the similarity distribution of foundation models. For an input $x$, the calibration loss is defined as
\begin{equation}
\mathcal{L}_{\mathrm{cal}}(x;\tau) 
= \mathrm{JS}\bigl(\widehat{P}_\theta(\cdot \mid x), \; P^{\mathrm{r}}_\tau(\cdot \mid x)\bigr),
\label{eq:calibration_loss_pointwise}
\end{equation}
where $P^{\mathrm{r}}_\tau(\cdot \!\mid\! x)$ is the temperature-scaled prediction of foundation model and $\widehat{P}_\theta(\cdot \!\mid\! x)$ is the prediction of the base model. The global calibration objective aggregates this loss over the unlabeled test distribution:
\begin{equation*}
\tau^\ast 
= \arg\min_{\tau>0} \;\mathbb{E}_{x \sim \mathcal{D}_{\mathrm{test}}}\bigl[\mathcal{L}_{\mathrm{cal}}(x;\tau)\bigr].
\label{eq:calibration_obj_population}
\end{equation*}
The temperature-scaled prediction of foundation model that appears in
\cref{eq:calibration_loss_pointwise}
is given by
\begin{equation*}
P^{\mathrm{r}}_\tau(j \mid x) 
= \frac{\exp\bigl(z^{\mathrm{r}}_j(x)/\tau\bigr)}
       {\sum_{k=1}^K \exp\bigl(z^{\mathrm{r}}_k(x)/\tau\bigr)},
\label{eq:temp_scaling}
\end{equation*}
where $z^{\mathrm{r}}_j(x)$ denotes the image-text cosine similarity computed by foundation model between input $x$ and the text embedding of class $j$. This test-time calibration procedure learns $\tau^\ast$ in a data-driven manner and adaptively mitigates under-confidence in foundation models. The complete pipeline is summarized as pseudo-code in \cref{alg:frap}.

\begin{algorithm}[h]
\caption{\FRAP(FRAP)}
\label{alg:frap}
\begin{algorithmic}[1]
    \STATE \textbf{Input:} base model $f_b$, foundation model $f_r$, labeled val dataset $\mathcal{D}_\text{val}$, unlabeled test dataset $\mathcal{D}_\text{test}$.
    \STATE \textbf{Output:} estimated error of base model on $\mathcal{D}_\text{test}$.
    \vspace{3pt}
    \STATE \textbf{(step 1) Test-time Calibration of $f_r$ prediction}
    \STATE Obtain $p_b(x)$ and $z_r(x)$ from $f_b$ and $f_r$, respectively.
    \WHILE{not converged}
        \STATE $\mathcal{L} \leftarrow \frac{1}{|\mathcal{B}|}\!\sum\limits_{x_i \in \mathcal{B}} \!\mathrm{JS}\big(p_b(x_i) \,\|\, \mathrm{softmax}(z_r(x_i)/\tau)\big)$
        \STATE $\tau \leftarrow \tau - \eta \cdot \nabla_\tau \mathcal{L}_{\mathcal{B}}$
    \ENDWHILE
    \STATE Calibration result $\tilde{p}_r(x) \leftarrow \mathrm{softmax}(z_r(x)/\tau^\ast)$.
    \vspace{3pt}
    \STATE \textbf{(step 2) Confidence-weighted Fusion}
    \STATE Fused prediction $\hat{p}(x)\leftarrow \alpha\,p_b(x)+(1-\alpha)\tilde{p}_r(x)$, where $\alpha$ is normalized confidence of $p_b(x)$ relative to $\tilde{p}_r(x)$.
    \vspace{3pt}
    \STATE \textbf{(step 3) Performance Estimation}
    \STATE Compute sample-wise score $s_i = \langle p_b(x_i),\, \hat{p}(x_i)\rangle$.
    \STATE Determine threshold $\delta$ from $\mathcal{D}_\text{val}$ and estimate error on $\mathcal{D}_\text{test}$ as the portion of samples satisfying $s_i \!\leq\! \delta$.
\end{algorithmic}
\end{algorithm}

\section{Experiments}
\label{sec: Exp}

In this section, we empirically evaluate FRAP and compare it against competitive baselines. We first train the base model on in-distribution source data with different random seeds and then evaluate the resulting checkpoints on unlabeled out-of-distribution data. The true labels of the test data are only used to compute the estimation error.

\noindent\textbf{Datasets.}
We evaluate our method on a diverse suite of 10 benchmark datasets covering both natural and synthetic distribution shifts, \ie, MNIST~\cite{LeCun1998GradientbasedLA}, CIFAR-10~\cite{Krizhevsky2009LearningML}, CIFAR-100~\cite{Krizhevsky2009LearningML}, ImageNet~\cite{Russakovsky2014ImageNetLS}, Tiny-ImageNet, FMoW~\cite{christie2018functional}, and four BREEDS datasets~\cite{santurkar2020breeds}. These benchmarks collectively span a wide range of real-world and subpopulation shifts, from digit recognition to large-scale object classification. For each source dataset, we consider its standard in-distribution training set and corresponding out-of-distribution test sets constructed from natural variants (\eg ImageNet-V2, CIFAR-10.1) and corruption benchmarks (\eg CIFAR-10/100-C, ImageNet-C). Further dataset specifications are provided in \cref{app:datasets}.

\noindent\textbf{Model Architectures.}
We consider different architectures tailored to each dataset. For MNIST, we construct a lightweight convolutional neural network. For CIFAR-10 and CIFAR-100, we utilize DenseNet121~\cite{Huang2016DenselyCC} and ResNet18~\cite{He2015DeepRL}. For ImageNet, Tiny-ImageNet, FMoW, and BREEDS, we adopt DenseNet121 and ResNet50~\cite{He2015DeepRL}. For each model, we train with three random seeds (\ie 0, 1, and 10) and save checkpoints at different epochs. Evaluation is conducted across architectures, random seeds, and training epochs to provide a comprehensive assessment of the estimation method. For the foundation model, we employ two vision-language models, CLIP with a ViT-B/32 backbone and SigLIP with a ViT-B/16 backbone, both using publicly available pre-trained weights without fine-tuning.

\begin{table*}[t]
\caption{\textbf{Comparison of FRAP and baselines.} Mean Absolute Error (MAE)(\%) $\downarrow$ across different methods and datasets. FRAP is evaluated using both CLIP and SigLIP as foundation models. Results reported by aggregating MAE numbers over different seeds and architectures, shown as \textbf{mean} (\%) \textbf{$\pm$ standard deviation}.}
\label{tab:main_results}
\vspace{-0.8em}
\begin{center}
\small
\setlength{\aboverulesep}{0pt} 
\setlength{\belowrulesep}{0pt} 
\setlength{\tabcolsep}{2pt} 
\resizebox{\linewidth}{!}{
\begin{tabular}{l|ccccccccc|cc}
\toprule
& \multicolumn{9}{c|}{\textbf{Baselines}} & \multicolumn{2}{c}{\textbf{Ours (FRAP)}} \\
\textbf{Dataset} & \textbf{IM} & \textbf{AC} & \textbf{DoC} & \textbf{GDE} & \textbf{ATC-MC} & \textbf{ATC-NE} & \textbf{projNorm} & \textbf{COT} & \textbf{COTT} & \textbf{SigLIP} & \textbf{CLIP} \\
\midrule
MNIST & 6.86{\scriptsize$\pm$.12} & 1.97{\scriptsize$\pm$.02} & \textbf{1.96}{\scriptsize$\pm$.02} & 25.05{\scriptsize$\pm$.27} & 10.27{\scriptsize$\pm$.10} & 13.80{\scriptsize$\pm$.11} & 14.77{\scriptsize$\pm$.18} & 6.04{\scriptsize$\pm$.06} & 13.81{\scriptsize$\pm$.09} & 10.56{\scriptsize$\pm$.10} & 12.42{\scriptsize$\pm$.10} \\
C10-N & 11.55{\scriptsize$\pm$.05} & 11.54{\scriptsize$\pm$.05} & 11.54{\scriptsize$\pm$.05} & 7.04{\scriptsize$\pm$.04} & 12.00{\scriptsize$\pm$.07} & 12.67{\scriptsize$\pm$.06} & 9.97{\scriptsize$\pm$.06} & 11.17{\scriptsize$\pm$.05} & 11.96{\scriptsize$\pm$.07} & 7.54{\scriptsize$\pm$.03} & \textbf{2.14}{\scriptsize$\pm$.02} \\
C10-S & 14.24{\scriptsize$\pm$.08} & 14.17{\scriptsize$\pm$.08} & 14.16{\scriptsize$\pm$.08} & 8.62{\scriptsize$\pm$.07} & 14.11{\scriptsize$\pm$.09} & 14.23{\scriptsize$\pm$.09} & 5.14{\scriptsize$\pm$.03} & 8.99{\scriptsize$\pm$.03} & 9.56{\scriptsize$\pm$.06} & 7.68{\scriptsize$\pm$.06} & \textbf{3.24}{\scriptsize$\pm$.05} \\
C100-N & 13.06{\scriptsize$\pm$.08} & 13.02{\scriptsize$\pm$.08} & 13.15{\scriptsize$\pm$.08} & 11.72{\scriptsize$\pm$.07} & 12.65{\scriptsize$\pm$.06} & 13.24{\scriptsize$\pm$.06} & 10.73{\scriptsize$\pm$.10} & 12.37{\scriptsize$\pm$.08} & 12.72{\scriptsize$\pm$.06} & 10.22{\scriptsize$\pm$.04} & \textbf{4.00}{\scriptsize$\pm$.03} \\
C100-S & 21.30{\scriptsize$\pm$.11} & 21.26{\scriptsize$\pm$.08} & 21.39{\scriptsize$\pm$.11} & 16.94{\scriptsize$\pm$.04} & 19.48{\scriptsize$\pm$.10} & 19.62{\scriptsize$\pm$.11} & 12.33{\scriptsize$\pm$.07} & 13.73{\scriptsize$\pm$.08} & 13.66{\scriptsize$\pm$.06} & 17.87{\scriptsize$\pm$.10} & \textbf{11.42}{\scriptsize$\pm$.10} \\
IN-S & 11.09{\scriptsize$\pm$.07} & 8.14{\scriptsize$\pm$.06} & 8.96{\scriptsize$\pm$.06} & 7.73{\scriptsize$\pm$.05} & 2.57{\scriptsize$\pm$.02} & 4.07{\scriptsize$\pm$.03} & 4.98{\scriptsize$\pm$.03} & 2.67{\scriptsize$\pm$.02} & \textbf{2.14}{\scriptsize$\pm$.02} & 3.74{\scriptsize$\pm$.02} & 4.76{\scriptsize$\pm$.01} \\
IN-N & 7.10{\scriptsize$\pm$.08} & 5.81{\scriptsize$\pm$.06} & 6.28{\scriptsize$\pm$.07} & 6.27{\scriptsize$\pm$.05} & 2.45{\scriptsize$\pm$.02} & 1.48{\scriptsize$\pm$.01} & 8.76{\scriptsize$\pm$.05} & 3.38{\scriptsize$\pm$.02} & 1.91{\scriptsize$\pm$.01} & \textbf{1.15}{\scriptsize$\pm$.01} & 2.16{\scriptsize$\pm$.03} \\
IN200-S & 11.53{\scriptsize$\pm$.09} & 7.99{\scriptsize$\pm$.08} & 9.06{\scriptsize$\pm$.08} & 30.22{\scriptsize$\pm$.13} & 3.36{\scriptsize$\pm$.05} & 6.86{\scriptsize$\pm$.04} & 4.96{\scriptsize$\pm$.04} & \textbf{3.02}{\scriptsize$\pm$.02} & 4.38{\scriptsize$\pm$.02} & 6.14{\scriptsize$\pm$.05} & 9.98{\scriptsize$\pm$.04} \\
IN200-N & 14.56{\scriptsize$\pm$.16} & 12.51{\scriptsize$\pm$.14} & 12.87{\scriptsize$\pm$.15} & 31.28{\scriptsize$\pm$.24} & 7.17{\scriptsize$\pm$.09} & 4.17{\scriptsize$\pm$.05} & 11.33{\scriptsize$\pm$.10} & 7.04{\scriptsize$\pm$.07} & \textbf{3.50}{\scriptsize$\pm$.05} & 5.43{\scriptsize$\pm$.07} & 7.46{\scriptsize$\pm$.04} \\
FMoW & 3.66{\scriptsize$\pm$.04} & 4.00{\scriptsize$\pm$.04} & 3.63{\scriptsize$\pm$.04} & 6.04{\scriptsize$\pm$.04} & 2.60{\scriptsize$\pm$.03} & \textbf{2.33}{\scriptsize$\pm$.03} & 22.24{\scriptsize$\pm$.10} & 8.02{\scriptsize$\pm$.05} & 8.76{\scriptsize$\pm$.06} & 2.70{\scriptsize$\pm$.03} & 3.42{\scriptsize$\pm$.02} \\
L17-S & 9.97{\scriptsize$\pm$.08} & 8.96{\scriptsize$\pm$.08} & 9.16{\scriptsize$\pm$.08} & 9.11{\scriptsize$\pm$.10} & 5.35{\scriptsize$\pm$.06} & 6.86{\scriptsize$\pm$.06} & 6.81{\scriptsize$\pm$.05} & \textbf{3.44}{\scriptsize$\pm$.03} & 4.09{\scriptsize$\pm$.03} & 6.81{\scriptsize$\pm$.06} & 8.55{\scriptsize$\pm$.05} \\
L17-N & 23.55{\scriptsize$\pm$.05} & 22.11{\scriptsize$\pm$.06} & 22.43{\scriptsize$\pm$.06} & 18.38{\scriptsize$\pm$.10} & 11.11{\scriptsize$\pm$.07} & 9.26{\scriptsize$\pm$.06} & 16.91{\scriptsize$\pm$.08} & 13.94{\scriptsize$\pm$.04} & 7.75{\scriptsize$\pm$.05} & 6.74{\scriptsize$\pm$.07} & \textbf{5.37}{\scriptsize$\pm$.05} \\
NL26-S & 11.63{\scriptsize$\pm$.11} & 9.54{\scriptsize$\pm$.10} & 10.01{\scriptsize$\pm$.10} & 12.63{\scriptsize$\pm$.14} & 6.87{\scriptsize$\pm$.09} & 7.98{\scriptsize$\pm$.08} & 6.54{\scriptsize$\pm$.05} & \textbf{3.18}{\scriptsize$\pm$.02} & 3.72{\scriptsize$\pm$.03} & 7.60{\scriptsize$\pm$.09} & 9.30{\scriptsize$\pm$.07} \\
NL26-N & 22.55{\scriptsize$\pm$.08} & 19.46{\scriptsize$\pm$.08} & 20.27{\scriptsize$\pm$.08} & 21.65{\scriptsize$\pm$.12} & 9.70{\scriptsize$\pm$.10} & 8.85{\scriptsize$\pm$.09} & 12.01{\scriptsize$\pm$.08} & 11.34{\scriptsize$\pm$.03} & \textbf{5.83}{\scriptsize$\pm$.04} & 8.48{\scriptsize$\pm$.10} & 6.66{\scriptsize$\pm$.09} \\
E13-S & 12.27{\scriptsize$\pm$.10} & 10.95{\scriptsize$\pm$.09} & 11.33{\scriptsize$\pm$.10} & 11.54{\scriptsize$\pm$.10} & 5.91{\scriptsize$\pm$.07} & 6.18{\scriptsize$\pm$.06} & 6.10{\scriptsize$\pm$.05} & 4.14{\scriptsize$\pm$.03} & \textbf{3.14}{\scriptsize$\pm$.03} & 5.78{\scriptsize$\pm$.07} & 8.29{\scriptsize$\pm$.05} \\
E13-N & 20.97{\scriptsize$\pm$.08} & 19.25{\scriptsize$\pm$.08} & 19.79{\scriptsize$\pm$.08} & 17.98{\scriptsize$\pm$.10} & 10.70{\scriptsize$\pm$.07} & 8.39{\scriptsize$\pm$.07} & 12.75{\scriptsize$\pm$.06} & 10.22{\scriptsize$\pm$.04} & \textbf{5.76}{\scriptsize$\pm$.03} & 8.50{\scriptsize$\pm$.07} & 5.90{\scriptsize$\pm$.06} \\
E30-S & 13.61{\scriptsize$\pm$.11} & 10.93{\scriptsize$\pm$.10} & 11.67{\scriptsize$\pm$.10} & 9.92{\scriptsize$\pm$.11} & 5.62{\scriptsize$\pm$.08} & 5.28{\scriptsize$\pm$.06} & 7.50{\scriptsize$\pm$.06} & 3.85{\scriptsize$\pm$.03} & \textbf{2.30}{\scriptsize$\pm$.02} & 5.81{\scriptsize$\pm$.08} & 6.94{\scriptsize$\pm$.06} \\
E30-N & 23.30{\scriptsize$\pm$.08} & 19.87{\scriptsize$\pm$.08} & 20.89{\scriptsize$\pm$.08} & 15.50{\scriptsize$\pm$.10} & 10.10{\scriptsize$\pm$.08} & 7.33{\scriptsize$\pm$.07} & 14.52{\scriptsize$\pm$.09} & 11.15{\scriptsize$\pm$.03} & 5.93{\scriptsize$\pm$.03} & 9.05{\scriptsize$\pm$.09} & \textbf{5.54}{\scriptsize$\pm$.07} \\
\midrule
\rowcolor{gray!20}
\textbf{Average} & 13.76 & 12.30 & 12.70 & 14.87 & 8.45 & 8.48 & 10.46 & 7.65 & 6.72 & 7.32 & \textbf{6.53} \\
\bottomrule
\end{tabular}
} 
\end{center}
\vspace{-1,3em}
\end{table*}

\subsection{Comparison Results}

We benchmark our method against several representative baselines for accuracy estimation under distribution shifts, including Importance Re-weighting (IM)~\cite{Chen2021MandolineME}, Average Confidence (AC), Difference of Confidence (DoC)~\cite{Guillory2021PredictingWC}, Generalized Disagreement Equality (GDE)~\cite{Jiang2021AssessingGO}, Average Thresholded Confidence (ATC)~\cite{Garg2022LeveragingUD}, including variants MC and NE, Projection Normalization (ProjNorm)~\cite{Yu2022PredictingOE,Lu2023CharacterizingOE}, Confidence Optimal Transport (COT) and its thresholded variant COTT~\cite{Lu2023CharacterizingOE}. Detailed formulations and implementation specifics for all baselines are provided in \cref{app:baselines}.

We evaluate all methods using Mean Absolute Estimation Error (MAE), defined as the average absolute difference between the true and estimated error rates on the target data. Experiments are conducted across multiple datasets, model architectures at different training epochs, and random seeds for a comprehensive comparison. As summarized in \cref{tab:main_results}, FRAP based on CLIP achieves the lowest average MAE of $6.53\%$, outperforming all baseline methods, with the next-best method COTT at $6.72\%$. 
While SigLIP yields a slightly higher average error ($7.32\%$), it remains highly competitive against specialized baselines like COT and ATC, demonstrating its robustness across different foundation models.
Furthermore, FRAP achieves the best result on 6 out of the 18 individual datasets, demonstrating superior and robust estimation performance. Regarding computational cost, COT/COTT scales quadratically or cubically with the number of classes due to the optimal transport computation. FRAP, in contrast, requires foundation model inference that scales linearly with the number of classes and avoids the expensive OT solver, making it substantially more efficient in high-cardinality label spaces.

\subsection{Calibration with Different Temperature}
\label{sec:Fixedtemp}

FRAP calibrates the predictions of foundation models using the dynamically learned temperature introduced in~\cref{sec:module1}. As shown in \cref{tab:ece_comparison}, our test-time calibration (TTC) effectively alleviates the miscalibration of the original similarity (Raw) in most settings. By reducing the Expected Calibration Error (ECE), TTC substantially narrows the gap between confidence scores and actual performance.
\begin{table}[t]
\caption{\textbf{Expected Calibration Error (ECE) $\downarrow$ comparison before and after TTC.} The results across various datasets demonstrate that TTC significantly reduces ECE for both \colorbox{pink!15}{CLIP} and \colorbox{yellow!10}{SigLIP} reference models.}
\vspace{-4mm}
\begin{center}
\tiny
\setlength{\tabcolsep}{1.5pt}
\resizebox{\linewidth}{!}{
\begin{tabular}{>{\columncolor{gray!15}}l>{\columncolor{gray!15}}l*{10}{c}} 
\toprule
\textbf{VLM} & \textbf{Method} & \textbf{C10} & \textbf{C100} & \textbf{FMoW} & \textbf{ImgN} & \textbf{MNIST} & \textbf{TinyImg} & \textbf{E13} & \textbf{E30} & \textbf{L17} & \textbf{NL26} \\
\midrule
\rowcolor{pink!15}
CLIP & TTC & \textbf{0.128} & \textbf{0.150} & \textbf{0.018} & \textbf{0.127} & 0.073 & \textbf{0.061} & \textbf{0.295} & \textbf{0.321} & \textbf{0.540} & \textbf{0.427} \\
\rowcolor{pink!15}
CLIP & Raw & 0.590 & 0.373 & 0.116 & 0.343 & \textbf{0.027} & 0.384 & 0.517 & 0.474 & 0.632 & 0.557 \\
\midrule
\rowcolor{yellow!10}
SigLIP & TTC & \textbf{0.159} & \textbf{0.209} & \textbf{0.034} & \textbf{0.171} & \textbf{0.283} & \textbf{0.092} & \textbf{0.281} & \textbf{0.348} & \textbf{0.419} & \textbf{0.422} \\
\rowcolor{yellow!10}
SigLIP & Raw & 0.632 & 0.478 & 0.122 & 0.427 & 0.510 & 0.456 & 0.501 & 0.499 & 0.668 & 0.567 \\
\bottomrule
\end{tabular}
}
\end{center}
\vspace{-1.8em}
\label{tab:ece_comparison}
\end{table}

\begin{figure}[t]
    \centering
    \includegraphics[width=\linewidth]{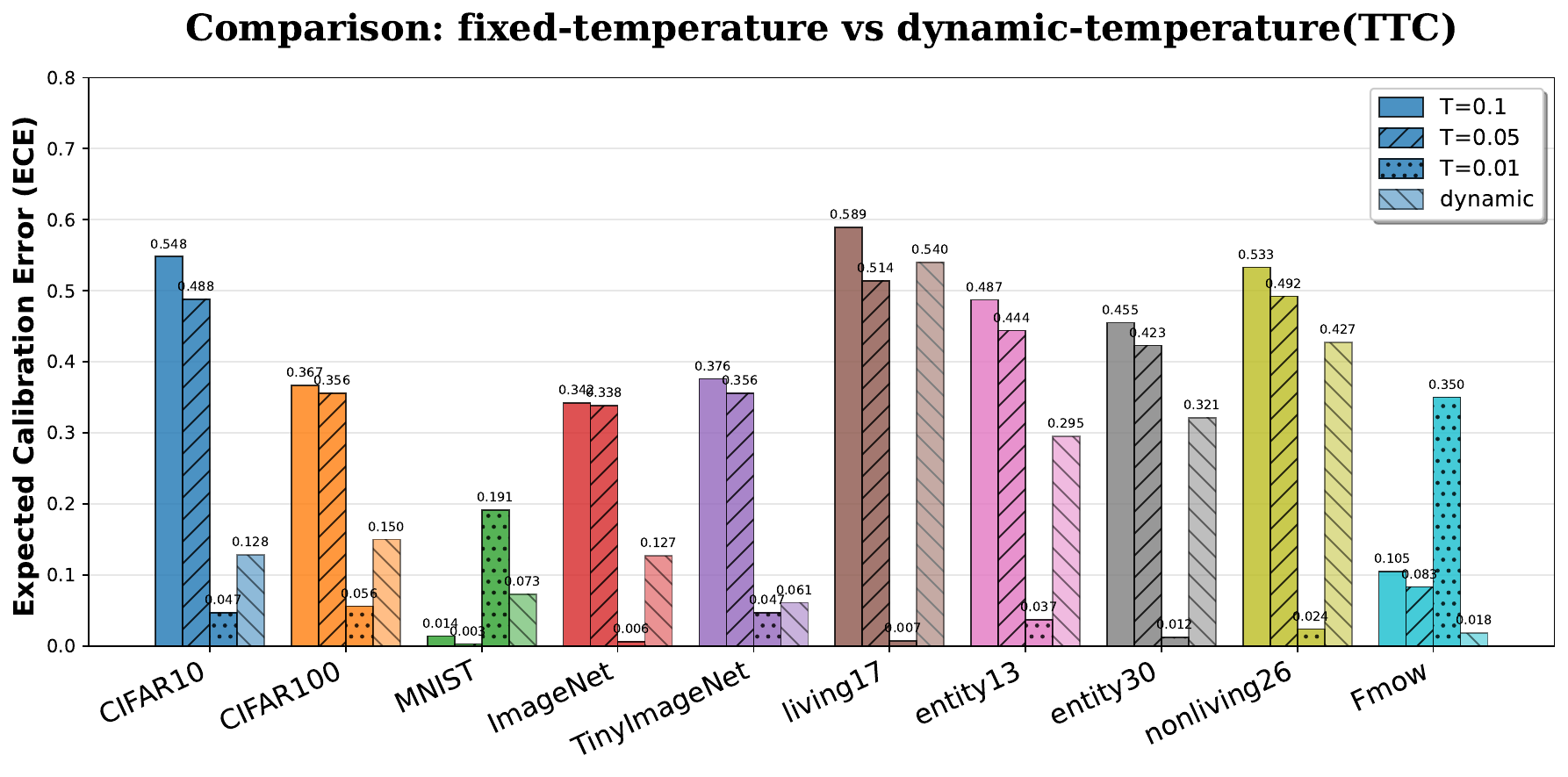}
    \vspace{-2em}
    \caption{\textbf{Fixed temperature \vs Test-time calibration.} We compare the ECE of our test-time-learned temperature (TTC) against the fixed temperature values ($\tau = 0.1, 0.05, 0.01$) across diverse datasets, with CLIP serving as the foundation model.
    }
    \label{fig:tempCompare}
    \vspace{-2em}
\end{figure}

We further explore the effect of fixed temperatures by adopting three constant values (\ie 0.1, 0.05, 0.01) and compare the calibration errors between our test-time-learned temperature and these fixed temperature coefficients. As shown in \cref{fig:tempCompare}, regarding the calibration error (ECE), our test-time-learned temperature (TTC) predominantly achieves superior results compared to fixed temperature scaling at $\tau=0.05$ and $\tau=0.1$, although the fixed small temperature $\tau=0.01$ yields the lowest calibration error in several cases. When the calibrated CLIP distributions are used for performance estimation, our standard FRAP with dynamically learned temperature significantly outperforms $\mathrm{FRAP}_{\tau=0.05}$ ($6.53\%$ \vs $8.19\%$), and marginally outperforms $\mathrm{FRAP}_{\tau=0.01}$ ($6.53\%$ \vs $6.61\%$). The full results are reported in \cref{tab:tempCompareResults}, where $\mathrm{RAP}$ denotes the method without temperature-scaling calibration, $\mathrm{FRAP}_{\tau=0.05}$ and $\mathrm{FRAP}_{\tau=0.01}$ denote calibration with fixed temperatures $0.05$ and $0.01$, respectively, and $\mathrm{FRAP}_{dyna}$ denotes our standard calibration with test-time-learned temperature. Although fixed small temperature $\tau = 0.01$ achieve lower calibration error, it leads to slightly worse estimation performance than TTC.

The results indicate that improved calibration, as measured by ECE, does not necessarily translate into improved estimation performance. We attribute this discrepancy to a fundamental mismatch between the calibration objective and the downstream estimation objective. Small fixed temperatures attain lower miscalibration error, while they are not optimized for our goal of enabling effective confidence-weighted fusion with the base model. The dynamically learned temperature, chosen to align the predictions of foundation model with those of the base model, implicitly acts as a regularizer that enforces compatible confidence scales between the two models and preserves the utility of the fusion scheme. This perspective explains why our approach, despite exhibiting higher miscalibration error than the $\tau=0.01$ variants, ultimately yields superior performance estimation. 

\begin{table}[t]
\caption{\textbf{Comparison of MAE under different temperature.} Results are reported by aggregating MAE numbers over different seeds and architectures, shown as \textbf{mean} (\%) \textbf{$\pm$ standard deviation}.}
\label{tab:tempCompareResults}
\vspace{-1em}
\begin{center}
\small
\setlength{\aboverulesep}{0pt} 
\setlength{\belowrulesep}{0pt} 
\setlength{\tabcolsep}{2.5pt}
\begin{tabular}{l|ccc|c}
\toprule
& \multicolumn{3}{c|}{\textbf{Fixed temperature}} & \multicolumn{1}{c}{\textbf{TTC}} \\
\textbf{Dataset} & \textbf{RAP} & \textbf{FRAP$_{\tau=0.05}$} & \textbf{FRAP$_{\tau=0.01}$} & \textbf{FRAP$_{dyna}$} \\
\midrule
MNIST & 26.31{\scriptsize$\pm$0.12} & 11.30{\scriptsize$\pm$0.10} & \textbf{11.13}{\scriptsize$\pm$0.10} & 12.42{\scriptsize$\pm$0.10} \\
C10-N & 7.26{\scriptsize$\pm$0.02} & 10.86{\scriptsize$\pm$0.04} & 8.48{\scriptsize$\pm$0.03} & \textbf{2.14}{\scriptsize$\pm$0.02} \\
C10-S & 7.47{\scriptsize$\pm$0.07} & 12.95{\scriptsize$\pm$0.07} & 8.01{\scriptsize$\pm$0.05} & \textbf{3.24}{\scriptsize$\pm$0.05} \\
C100-N & 12.24{\scriptsize$\pm$0.06} & 12.57{\scriptsize$\pm$0.06} & 10.90{\scriptsize$\pm$0.05} & \textbf{4.00}{\scriptsize$\pm$0.03} \\
C100-S & \textbf{7.15}{\scriptsize$\pm$0.06} & 19.56{\scriptsize$\pm$0.10} & 17.07{\scriptsize$\pm$0.10} & 11.42{\scriptsize$\pm$0.10} \\
IN-S & 6.93{\scriptsize$\pm$0.03} & \textbf{2.94}{\scriptsize$\pm$0.02} & 3.11{\scriptsize$\pm$0.02} & 4.76{\scriptsize$\pm$0.01} \\
IN-N & 1.61{\scriptsize$\pm$0.01} & 1.99{\scriptsize$\pm$0.02} & \textbf{1.43}{\scriptsize$\pm$0.01} & 2.16{\scriptsize$\pm$0.03} \\
IN200-S & 16.20{\scriptsize$\pm$0.07} & \textbf{4.34}{\scriptsize$\pm$0.04} & 4.95{\scriptsize$\pm$0.04} & 9.98{\scriptsize$\pm$0.04} \\
IN200-N & \textbf{2.85}{\scriptsize$\pm$0.02} & 6.12{\scriptsize$\pm$0.07} & 5.32{\scriptsize$\pm$0.06} & 7.46{\scriptsize$\pm$0.04} \\
FMoW & 8.10{\scriptsize$\pm$0.05} & \textbf{2.51}{\scriptsize$\pm$0.03} & 3.39{\scriptsize$\pm$0.03} & 3.42{\scriptsize$\pm$0.02} \\
L17-S & 11.71{\scriptsize$\pm$0.07} & \textbf{5.74}{\scriptsize$\pm$0.06} & 6.15{\scriptsize$\pm$0.05} & 8.55{\scriptsize$\pm$0.05} \\
L17-N & 6.92{\scriptsize$\pm$0.04} & 10.57{\scriptsize$\pm$0.07} & 5.70{\scriptsize$\pm$0.04} & \textbf{5.37}{\scriptsize$\pm$0.05} \\
NL26-S & 6.76{\scriptsize$\pm$0.03} & 7.20{\scriptsize$\pm$0.09} & \textbf{5.64}{\scriptsize$\pm$0.05} & 9.30{\scriptsize$\pm$0.07} \\
NL26-N & \textbf{4.61}{\scriptsize$\pm$0.03} & 9.14{\scriptsize$\pm$0.09} & 5.74{\scriptsize$\pm$0.05} & 6.66{\scriptsize$\pm$0.09} \\
E13-S & 5.11{\scriptsize$\pm$0.04} & 5.73{\scriptsize$\pm$0.06} & \textbf{4.93}{\scriptsize$\pm$0.05} & 8.29{\scriptsize$\pm$0.05} \\
E13-N & 6.92{\scriptsize$\pm$0.04} & 9.43{\scriptsize$\pm$0.07} & 6.98{\scriptsize$\pm$0.06} & \textbf{5.90}{\scriptsize$\pm$0.06} \\
E30-S & 4.69{\scriptsize$\pm$0.03} & 5.42{\scriptsize$\pm$0.07} & \textbf{4.07}{\scriptsize$\pm$0.05} & 6.94{\scriptsize$\pm$0.06} \\
E30-N & \textbf{4.74}{\scriptsize$\pm$0.03} & 9.05{\scriptsize$\pm$0.08} & 6.01{\scriptsize$\pm$0.05} & 5.54{\scriptsize$\pm$0.07} \\
\midrule
\rowcolor{gray!20}
\textbf{Average} & 8.20 & 8.19 & 6.61 & \textbf{6.53}\\
\bottomrule
\end{tabular}
\end{center}
\vspace{-1.5em}
\end{table}

\subsection{Effectiveness of Confidence-Weighted Fusion}
\label{sec:VerifyCWF}

\begin{figure*}[t]
    \centering
    \begin{subfigure}[b]{\columnwidth}
        \centering
        \includegraphics[width=\textwidth]{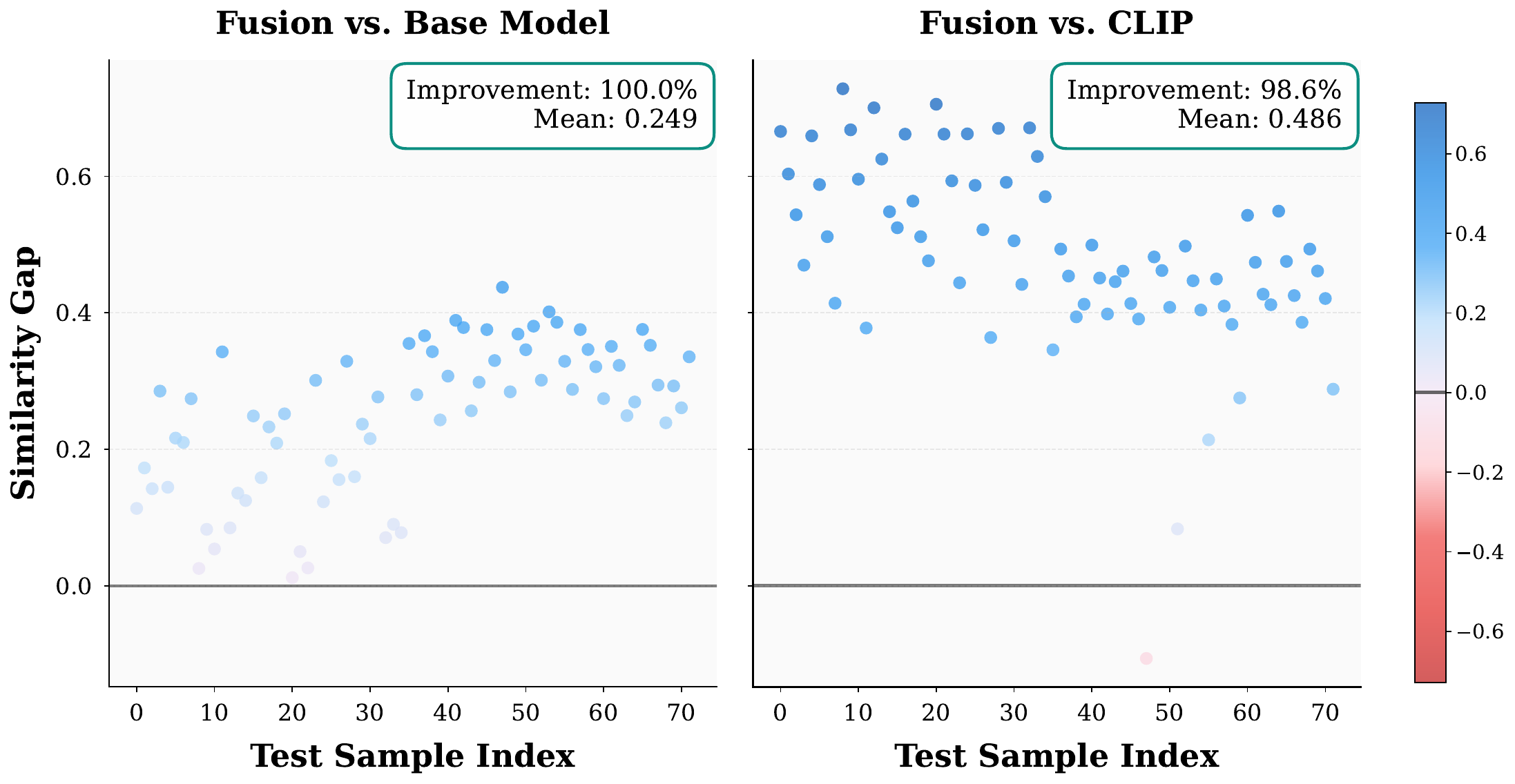}
        \caption{Natural shift}
        \label{fig:NaturalShiftGap}
    \end{subfigure}
    \hfill
    \begin{subfigure}[b]{\columnwidth}
        \centering
        \includegraphics[width=\textwidth]{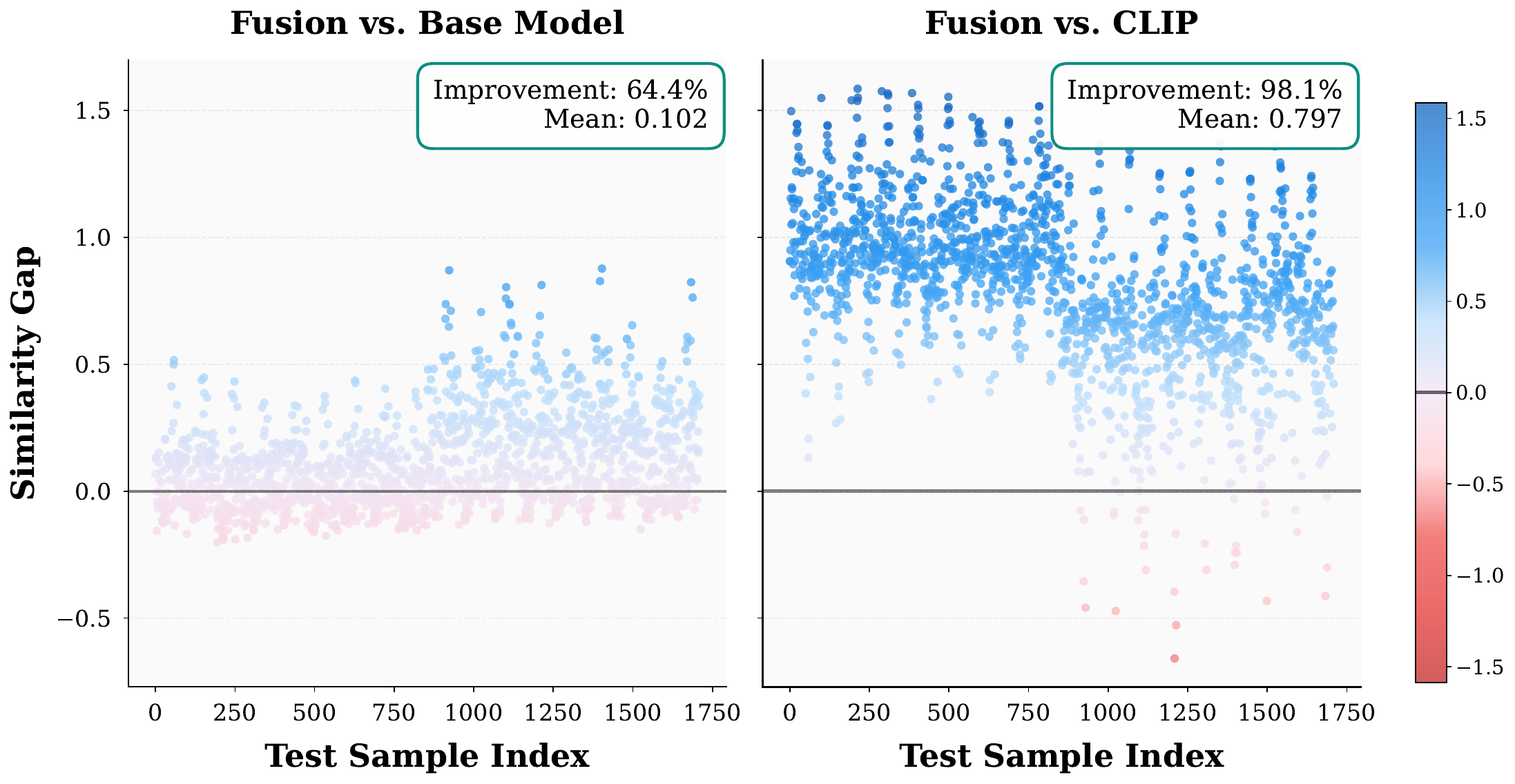}
        \caption{Synthetic shift}
        \label{fig:SyntheticShiftGap}
    \end{subfigure}
    \caption{\textbf{Effectiveness of CWF.} Per-experiment \SAS (SAS) gaps on ImageNet. We compare the confidence-weighted fusion against the base model (left column) and CLIP (right column) under (a) natural shifts and (b) synthetic corruptions. Positive similarity gap $\Delta\text{SAS} = \text{SAS}_{\text{fusion}} - \text{SAS}_{\text{baseline}}$ indicate improved semantic alignment ($\Delta\text{SAS} > 0$) and the fusion predominantly achieves positive differences, validating its effectiveness in preserving semantic coherence.}
    \label{fig:SimilarityGap}
    \vspace{-0.5em}
\end{figure*}

To further assess the effectiveness of the confidence-weighted fusion strategy, we introduce the \SAS (SAS), which measures the semantic consistency between predictions and ground-truth labels. This analysis is conducted on several datasets constructed based on WordNet~\cite{Miller1995WordNetAL}, a human-curated lexical database that organizes concepts into a tree-structured taxonomy. Following~\cite{Rada1989DevelopmentAA, Wu1994VerbSA}, the semantic proximity of two concepts can be measured by \textit{taxonomic distance} $d(\cdot, \cdot)$ denoting the shortest-path edge count between them, \textit{path similarity} $s_{\text{path}}(\cdot, \cdot)$ as the inverse of $d(\cdot, \cdot)$, and \textit{Wu-Palmer similarity} $s_{\text{wup}}(\cdot, \cdot)$ based on the depth of their lowest common ancestor relative to the depths of the individual concepts. We then compute the comprehensive semantic similarity between $c_i$ and $c_j$ as
\begin{equation}
\label{eq:classSimilarity}
    \mathcal{S}(c_i, c_j) = s_{\text{wup}}(c_i, c_j) + s_{\text{path}}(c_i, c_j) - d(c_i, c_j).
\end{equation}

Given a predicted probability distribution $\mathbf{p} = [p_1, \ldots, p_K] \in \mathbb{R}^K$ over $K$ classes and the ground-truth label $y \in \{1, \ldots, K\}$, SAS is defined as
\begin{equation*}
\label{eq:sas}
    \text{SAS}(\mathbf{p}, y) = \sum_{i=1}^{K} p_i \cdot \mathcal{S}(c_i, c_y),
\end{equation*}
where $p_i$ is the predicted probability for class $c_i$, and $\mathcal{S}(c_i, c_y)$ measures the semantic similarity between class $c_i$ and the ground-truth class $c_y$ as in \cref{eq:classSimilarity}. Higher SAS indicates better semantic alignment with the true label, even when the top-1 prediction is incorrect. 

As shown in~\cref{fig:SimilarityGap}, we compute the SAS gap $\Delta\text{SAS} = \text{SAS}_{\text{fusion}} - \text{SAS}_{\text{baseline}}$ on ImageNet and its variants, where the baseline is either the base model or CLIP. Under natural distribution shifts, illustrated in~\cref{fig:NaturalShiftGap}, confidence-weighted fusion consistently achieves positive $\Delta\text{SAS}$ values across nearly all experiments. Under synthetic corruptions, illustrated in~\cref{fig:SyntheticShiftGap}, fusion maintains predominantly positive differences across the majority of test cases, indicating that the proposed strategy produces predictions that are more semantically coherent with ground-truth labels. 

We conduct the same analysis on additional datasets including Tiny-ImageNet, Living-17, and Nonliving-26. The results show that confidence-weighted fusion generally improves semantic alignment, with particularly notable gains on Living-17 and Nonliving-26. Detailed results and analysis are provided in \cref{app:AdditionalCWF}.

\subsection{Thresholding: Bridge the Gap to the Ideal}
\label{sec:Thresholding}

The fused predictions provide a practical surrogate in the absence of ground truth, and this approximation is inherently imperfect and inevitably deviates from the ideal one-hot labels, which leads to non-negligible estimation errors. To compensate for this gap, we introduce a simple yet effective thresholding strategy inspired by~\cite{Garg2022LeveragingUD}.

Concretely, let $\mathrm{Est}(x)$ denote the estimation score that serves as an approximation of the accuracy, computed as in \cref{eq:root}, where the ideal label distribution is approximated by the fusion of the two predictive models. We define a threshold $\delta$ on the source validation set $\mathcal{D}_s$ such that the fraction of samples with $\mathrm{Est}(x)$ below $t$ matches the error rate of $f_\theta$:
\begin{equation*}
    \frac{1}{|\mathcal{D}_s|}\! \sum_{x \in \mathcal{D}_s} \mathbb{I}\{\mathrm{Est}(x)\! < \!\delta\} 
    = \frac{1}{|\mathcal{D}_s|}\! \sum_{(x,y) \in \mathcal{D}_s}\!\mathbb{I}\{\widehat{y}(x)\!\neq \!y\},
\end{equation*}
where $\widehat{y}(x) = \arg \max_j \widehat{P}_\theta(j \!\mid\! x)$ and $\widehat{P}_\theta(\cdot \!\mid\! x)$ is the predictive distribution of the base model $f_{\theta}$. This strategy effectively reduces the problem from estimating absolute score values to making binary correctness decisions. Rather than attempting to perfectly approximate the ideal label distribution, the estimator only needs to decide whether the score lies above or below the threshold. Since $\mathrm{Est}(\cdot)$ is derived from the fusion of the two predictive models, it inherently combines their complementary strengths. By applying the same threshold $\delta$ to the unlabeled test set, the proportion of samples with $\mathrm{Est}(x) < \delta$ yields the estimated error rate of the classifier under distribution shift in our method.

\subsection{Reference Model Dependency and Robustness}

\begin{table}[h]
\vspace{-2mm}
\caption{\textbf{Extensive baselines.} MAE$\downarrow$ on {\setlength{\fboxsep}{1.5pt}\colorbox{pink!15}{CLIP}}, {\setlength{\fboxsep}{1.5pt}\colorbox{yellow!15}{SigLIP}}, and {\setlength{\fboxsep}{1.5pt}\colorbox{green!10}{Random}} reference for \textbf{Base} (pseudo-labels from the reference), \textbf{Fix} (FRAP with fixed scaling), \textbf{TTC} (FRAP with Test-time-calibration), and \textbf{Random} (FRAP with a random reference).}
\vspace{-4mm}
\label{tab:comparison}
\begin{center}
\setlength{\tabcolsep}{5pt}
\resizebox{\linewidth}{!}{
\begin{tabular}{>{\columncolor{gray!15}}l*{3}{>{\columncolor{pink!15}}c}*{3}{>{\columncolor{yellow!10}}c}>{\columncolor{green!10}}c} 
\toprule
 & \textbf{Base} & \textbf{Fix} & \textbf{TTC} & \textbf{Base} & \textbf{Fix} & \textbf{TTC} & \textbf{Random} \\
\midrule
\textbf{MAE(avg)} & 10.32 & 6.61 & \textbf{6.53} & 9.21 & \textbf{5.65} & 7.32 & 8.10 \\
\bottomrule
\end{tabular}
}
\end{center}
\vspace{-2em}
\label{tab:reference_clarify}
\end{table}

To further investigate the source of performance gains and the robustness of our framework, we conduct an extensive analysis across different reference sources.

\noindent\textbf{Methodological Gain \vs External Model Benefits.} 
A natural concern is whether the performance improvement of FRAP stems solely from the inclusion of powerful foundation models. We introduce a baseline (\textbf{Base}) that directly employs the pseudo-labels from the reference model as the ground-truth for accuracy estimation. As shown in \cref{tab:reference_clarify}, FRAP with test-time calibration (\textbf{Dyna}) and its fixed-temperature calibration variant (\textbf{Fix}) consistently outperform the \textbf{Base} method under various benchmark settings (CLIP and SigLIP). This gap demonstrates that FRAP benefits from our proposed framework rather than merely relying on the zero-shot capability of reference model.

\noindent\textbf{Robustness under Reference Failure.} 
To test the robustness of FRAP for reference model, we conduct a stress test using a \textbf{Random} reference generated via a Dirichlet distribution. As reported in \cref{tab:reference_clarify}, FRAP degrades gracefully under this pathological setting instead of failing catastrophically. This demonstrates the architectural-agnostic robustness even when the reference quality is severely compromised. 

\noindent\textbf{TTC for Reference-Agnostic Reliability.} 
As demonstrated in \cref{tab:reference_clarify}, foundation models like CLIP and SigLIP exhibit impressive performance when their internal scaling parameters (\eg, $\tau\approx0.01$ for CLIP ViT-B/32) are applied (\textbf{Fix}). In particular, Fix (SigLIP) reaches a new SOTA in estimation accuracy across several benchmarks. 
However, we argue that our proposed Test-Time Calibration (\textbf{TTC}) remains essential for the FRAP framework. One fundamental value of TTC lies in removal of fragile reliance on perfectly calibrated reference models, enabling reliable estimation even when the reference is miscalibrated, or entirely broken. Our Random stress tests confirm that TTC is indispensable when employing smaller, non-foundation networks as references to save computational costs.

\noindent\textbf{Selection of Reference Model.} 
While our framework is reference-agnostic, the choice of reference inevitably influences the estimation ceiling. How to systematically select the most suitable reference model remains an open question, which we leave for future research.

\section{Conclusion}

This work introduces \FRAP (FRAP), a framework for estimating model performance on unlabeled test data under distribution shift by jointly exploiting a task-specialized base model and a broadly trained foundation model. FRAP first reformulates accuracy as an average inner product between the base model predictions and the underlying label distributions, and then approximates the latter with a fused reference distribution. This reference is obtained by test-time calibration of the foundation model via Jensen–Shannon divergence minimization, followed by confidence-weighted fusion that integrates cross-domain generalization from the foundation model with domain-specific expertise from the base model. Combined with a simple thresholding scheme, FRAP achieves average lower mean absolute estimation error than representative baselines across diverse benchmark settings, and remains computationally efficient in high-cardinality label spaces.

There are several limitations and directions for further investigation of our work. FRAP does not consistently outperform baselines across all benchmarks as the effectiveness correlates with the generalization limits of the foundation model. These observations point to promising avenues for future work, including developing domain-adapted or task-specific reference models, extending FRAP to other modalities such as natural language via strong language foundations, and designing fusion and calibration objectives that are more tightly coupled to the performance-estimation goal.

\section*{Acknowledgments}
This work was supported partially by the NSFC General Program (62576367), the NSFC Excellent Young Scientists Fund (Overseas) (2025HY00260105, 2025HYSPT0708), and the Scientific Research Startup Fund at Sun Yat-sen University (25hytd012).

{
    \small
    \bibliographystyle{ieeenat_fullname}
    \bibliography{main}
}


\clearpage
\setcounter{page}{1}
\maketitlesupplementary
\appendix
\numberwithin{equation}{section}

\section{Derivation of Proposition~1}
\label{app:proof}

For a finite target sample set $\{(x_i, y_i)\}_{i=1}^N$, the standard empirical accuracy (\ie, Hard Accuracy) of the model $f_\theta$ is defined as:
\begin{equation}
    \mathrm{ACC}_N = \frac{1}{N}\sum_{i=1}^N \mathbb{I}\{\widehat{y}(x_i) = y_i\},
\end{equation}
where $\widehat{y}(x) = \arg\max_{j} \widehat{P}_\theta(j \mid x)$ is the hard label prediction. Since the hard indicator function is non-differentiable and difficult to analyze directly, we consider the expected soft accuracy as a surrogate for theoretical analysis. The soft accuracy replaces the hard match with the probability of the true class:
\begin{equation}
    \mathrm{ACC}^{\text{soft}}_N = \frac{1}{N}\sum_{i=1}^N \widehat{P}_\theta(y_i \mid x_i).
\end{equation}

By the Law of Large Numbers (LLN), as $N \to \infty$, the empirical soft accuracy converges to its population expectation over the target distribution $\mathcal{D}_t$:
\begin{equation}
    \lim_{N \to \infty} \mathrm{ACC}^{\text{soft}}_N 
    = \mathbb{E}_{(X, Y) \sim \mathcal{D}_t} \big[ \widehat{P}_\theta(Y \mid X) \big]
    \;\triangleq\;
    \mathbb{E}[\mathrm{ACC}_{\text{soft}}].
\end{equation}

Using the law of iterated expectations (conditioning on $X$), we expand this term as:
\begin{equation}
\label{eq:derivation}
    \mathbb{E}[\mathrm{ACC}_{\text{soft}}] = \int_{\mathcal{X}} \mathbb{E}_{Y \mid x} \big[ \widehat{P}_\theta(Y \mid x) \big] \, p_t(x) \, \mathrm{d}x.
\end{equation}

Inside the integral, the term $\mathbb{E}_{Y \mid x} [ \widehat{P}_\theta(Y \mid x) ]$ represents the expected confidence score of the model with respect to the label variable $Y$ given $X=x$. By the definition of expectation for a discrete random variable, this expectation expands to the sum of the function values weighted by their probabilities:
\begin{equation}
    \mathbb{E}_{Y \mid x} \big[ \widehat{P}_\theta(Y \mid x) \big] = \sum_{j=1}^K \underbrace{P^\ast(j \mid x)}_{\text{True Probability}} \cdot \underbrace{\widehat{P}_\theta(j \mid x)}_{\text{Model Prediction}},
\end{equation}
where $P^\ast(j \mid x) \triangleq \mathbb{P}(Y=j \mid X=x)$ denotes the ground-truth class posterior, and $\widehat{P}_\theta(j \mid x)$ denotes the predictive probability of the model for class $j$.

Substituting this expansion back into \cref{eq:derivation}, we obtain:
\begin{equation}
    \mathbb{E}[\mathrm{ACC}_{\text{soft}}]
    = \int_{\mathcal{X}} \left( \sum_{j=1}^K \widehat{P}_\theta(j \mid x)\, P^\ast(j \mid x) \right) p_t(x)\,\mathrm{d}x.
\end{equation}

Finally, approximating the outer integral with the finite target samples $\{x_i\}_{i=1}^N$ via Monte Carlo estimation yields:
\begin{equation}
\label{appeq:root}
    \mathbb{E}[\mathrm{ACC}_{\text{soft}}] \;\approx\; \frac{1}{N}\sum_{i=1}^N \sum_{j=1}^K \widehat{P}_\theta(j \mid x_i)\, P^\ast(j \mid x_i).
\end{equation}

\section{Contrastive representation learning}
\label{app:contrastiveLearning}

Many contrastive embedding models are trained using the InfoNCE (Information Noise-Contrastive Estimation) objective~\cite{Oord2018RepresentationLW,Chen2020ASF,Radford2021LearningTV,Wang2020UnderstandingCR,Robinson2021CanCL}. Fundamentally, this loss shapes the embedding space by pulling representations of positive (matched) pairs closer while pushing negative (unmatched) pairs apart.

Formally, given a mini-batch of $N$ samples, for each query sample $q_i$, the InfoNCE loss is defined as:
\begin{equation}
\label{appeq:contrastiveLoss}
\mathcal{L}_{\text{InfoNCE}} = -\frac{1}{N}\sum_{i=1}^{N} \log \frac{\exp(\text{sim}(\mathbf{q}_i, \mathbf{k}_i^+) / \tau)}{\sum_{j=1}^{N} \exp(\text{sim}(\mathbf{q}_i, \mathbf{k}_j) / \tau)},
\end{equation}
where $\mathbf{q}_i, \mathbf{k}_i^+ \in \mathbb{R}^d$ denote the embeddings of the query and its corresponding positive key, and $\{\mathbf{k}_j\}_{j=1}^N$ includes the positive key $\mathbf{k}_i^+$ and $N-1$ negative keys. The function $\text{sim}(\mathbf{u}, \mathbf{v}) = \mathbf{u}^\top \mathbf{v} / (\|\mathbf{u}\| \|\mathbf{v}\|)$ computes cosine similarity, and $\tau \in \mathbb{R}^+$ is a temperature parameter controlling the concentration of the distribution.

\subsection{CLIP Training}
CLIP applies InfoNCE to learn aligned image-text representations. Given a batch of $N$ image-text pairs $\{(I_i, T_i)\}_{i=1}^{N}$, CLIP encodes images and texts into $\ell_2$-normalized embeddings $\mathbf{z}_i^v, \mathbf{z}_i^t \in \mathbb{R}^d$ using vision encoder $f_v$ and text encoder $f_t$ respectively:
\begin{equation}
\mathbf{z}_i^v = \frac{f_v(I_i)}{\|f_v(I_i)\|_2}, \quad \mathbf{z}_i^t = \frac{f_t(T_i)}{\|f_t(T_i)\|_2}.
\end{equation}

The symmetric contrastive loss is calculated as the average of the image-to-text ($\mathcal{L}^{I \rightarrow T}$) and text-to-image ($\mathcal{L}^{T \rightarrow I}$) losses. For the $i$-th sample, the image-to-text loss is:
\begin{equation}
    \mathcal{L}_i^{I \rightarrow T} = -\log \frac{\exp(\text{sim}(\mathbf{z}_i^v, \mathbf{z}_i^t)/\tau)}{\sum_{j=1}^{N} \exp(\text{sim}(\mathbf{z}_i^v, \mathbf{z}_j^t)/\tau)},
\end{equation}
where $\text{sim}(\mathbf{z}_i^v, \mathbf{z}_j^t) = (\mathbf{z}_i^v)^\top \mathbf{z}_j^t$ is the cosine similarity. Symmetrically, the text-to-image loss is:
\begin{equation}
    \mathcal{L}_i^{T \rightarrow I} = -\log \frac{\exp(\text{sim}(\mathbf{z}_i^t, \mathbf{z}_i^v)/\tau)}{\sum_{j=1}^{N} \exp(\text{sim}(\mathbf{z}_i^t, \mathbf{z}_j^v)/\tau)}.
\end{equation}

The final CLIP loss is the average of both directions:
\begin{equation}
    \mathcal{L}_{\text{CLIP}} = \frac{1}{2N}\sum_{i=1}^{N} \left(\mathcal{L}_i^{I \rightarrow T} + \mathcal{L}_i^{T \rightarrow I}\right),
\end{equation}

\subsection{Mechanism of Uniformity}
To understand why CLIP yields predictions with high uniformity (\ie, over-smoothed distributions), we analyze the InfoNCE loss through two complementary lenses: \textit{Alignment-Uniformity decomposition} and \textit{Gradient dynamics}.

\paragraph{Alignment and Uniformity}
Previous work~\cite{Wang2020UnderstandingCR} decomposes the quality of learned representations into two key properties:
\begin{itemize}
    \item \textbf{Alignment} measures the closeness of positive pairs.
    \item \textbf{Uniformity} quantifies how uniformly features are distributed on the hyper-sphere.
\end{itemize}
InfoNCE implicitly optimizes both objectives, but the balance between them evolves during training. In the early stage (\ie Alignment phase), the model rapidly pulls positive pairs closer and the numerator term in InfoNCE dominates optimization, as positive similarities are initially low. In the late stage (\ie Uniformity phase), once positive pairs are sufficiently aligned, the numerator approaches its maximum and the gradient signal from the numerator nearly diminishes. Thus the optimization shifts to minimizing the denominator of the InfoNCE loss, which encourages spreading out negative pairs uniformly across the hypersphere~\cite{Schrodi2024TwoEO}. Consequently, the resulting prediction probabilities tend to spread out over all classes rather than concentrating on a single peak, especially when semantic ambiguity exists.

\paragraph{Implicit Hard Negative Mining}
Another explanation takes a different but consistent perspective that the InfoNCE performs implicit hard negative mining via its exponential weighting scheme~\cite{Dinu2025EffectivePE}: 
\begin{equation}
\frac{\partial \mathcal{L}_{\text{InfoNCE}}}{\partial \text{sim}(\mathbf{q}_i, \mathbf{k}_j)} \propto \exp(\text{sim}(\mathbf{q}_i, \mathbf{k}_j)/\tau), \quad j \neq i,
\end{equation}
which means that hard negatives receive exponentially larger gradients, causing the model to focus more on distinguishing confusing negative samples. Consequently, InfoNCE automatically emphasizes learning from the most challenging examples without explicit hard negative mining.

Let us derive the gradient of InfoNCE loss, which is an equivalent form of the formulation~\cref{appeq:contrastiveLoss}:
\begin{equation}
    \mathcal{L}_i = -\log \frac{\exp(s_i^+ / \tau)}{\exp(s_i^+ / \tau) + \sum_{j \neq i} \exp(s_j^- / \tau)},
\end{equation}
where $s_i^+$ denotes the positive pair similarity and $s_j^-$ denotes the negative pair similarity. For a negative pair similarity $s_j^-$, the gradient is:
\begin{align}
    \frac{\partial \mathcal{L}_i}{\partial s_j^-}
    &= \frac{\partial}{\partial s_j^-}
       \left[-\log
       \frac{\exp(s_i^+ / \tau)}
       {\exp(s_i^+ / \tau) + \sum_{k \neq i} \exp(s_k^- / \tau)}\right] \notag\\[4pt]
    &= \frac{\partial}{\partial s_j^-} \left[
       \log\!\left(
       \exp(s_i^+ / \tau) + \sum_{k \neq i} \exp(s_k^- / \tau)\right) \!
       - \! \frac{s_i^+}{\tau}. \notag \right] \\[4pt]
    &= \frac{1}{\tau}
       \frac{\exp(s_j^- / \tau)}
       {\exp(s_i^+ / \tau) + \sum_{k \neq i} \exp(s_k^- / \tau)}.
\end{align}

Therefore, the gradient magnitude is proportional to the exponential term:
\begin{equation}
    \frac{\partial \mathcal{L}_i}{\partial s_j^-}
    \propto \exp(s_j^- / \tau),
\end{equation}
which shows that negative samples with higher similarity (hard negatives) receive larger gradients. This implicit weighting mechanism naturally emphasizes hard negatives during optimization, leading to more uniform feature distribution to separate these challenging cases on the hypersphere, as discussed in~\cite{Dinu2025EffectivePE,Schrodi2024TwoEO}. This training dynamic is fundamentally intertwined with the temperature parameter $\tau$. Since CLIP trains a learnable temperature which typically converges to a very low value ($\tau \approx 0.01$), this low $\tau$ imposes severe uniformity pressure on the feature space by aggressively magnifying the gradient signal from hard negatives. Though low $\tau$ mathematically causes the softmax output to be sharper (high confidence) by magnifying logit differences, the resultant high feature uniformity is the root cause of the observed calibration issue: it leads to overly uniform prediction distributions for inputs lacking strong semantic alignment, as we observe in CLIP.

\section{Datasets}
\label{app:datasets}

\begin{table*}[t]
\caption{Summary of datasets and their corresponding distribution shifts. For CIFAR, ImageNet, Tiny-ImageNet and MNIST, the suffixes `-N' and `-S' denote \textit{Natural} and \textit{Synthetic} distribution shifts, respectively. For BREEDS, `-S' and `-N' denote \textit{Same} and \textit{Novel} subpopulations. BREEDS-S consists of L17-S, NL26-S, E13-S, and E30-S; BREEDS-N consists of L17-N, NL26-N, E13-N, and E30-N.}
\begin{center}
\small
\setlength{\tabcolsep}{12pt}
\begin{tabular}{p{2cm}p{2cm}p{10cm}}
\toprule
\textbf{Dataset} & \textbf{Shift Type} & \textbf{Shift Dataset(s)} \\
\midrule
MNIST & --  & QMNIST, USPS, SVHN \\
\midrule
\multirow{2}{*}{CIFAR-10} & C10-N & CIFAR-10v2 (re-sampled natural test set) \\
& C10-S & CIFAR-10-C (19 corruption types × 5 severity levels = 95 corruptions) \\
\midrule
\multirow{2}{*}{CIFAR-100} & C100-N & CIFAR-100 test set (standard evaluation) \\
& C100-S & CIFAR-100-C (19 corruption types × 5 severity levels = 95 corruptions) \\
\midrule
\multirow{3}{*}{ImageNet} & \multirow{2}{*}{IN-N} & ImageNet-V2 (3 variants: Matched-Frequency, Threshold-0.7, Top-Images), \\ & & ImageNet-Sketch \\
& IN-S & ImageNet-C (19 corruption types × 5 severity levels = 95 corruptions)\\
\midrule
\multirow{2}{*}{Tiny-ImageNet} & IN200-N & Imagenet200-V2 (3 variants), Imagenet200-Sketch, ImageNet-Reality \\
& IN200-S & ImageNet200-C (19 corruption types × 5 severity levels = 95 corruptions) \\
\midrule
FMoW & -- & FMoW (OOD-val and OOD-test) \\
\midrule
\multirow{2}{*}{BREEDS} & BREEDS-S & IN-N and IN-S with same subpopulation hierarchies \\
& BREEDS-N & IN-N and IN-S with novel subpopulation hierarchies \\
\bottomrule
\end{tabular}
\label{tab:datasets}
\end{center}
\end{table*}

We focus on natural image classification and evaluate our method on 10 benchmark datasets encompassing both natural and synthetic distribution shifts. Specifically, we use MNIST~\cite{LeCun1998GradientbasedLA}, CIFAR-10~\cite{Krizhevsky2009LearningML}, CIFAR-100~\cite{Krizhevsky2009LearningML}, ImageNet~\cite{Russakovsky2014ImageNetLS}, Tiny-ImageNet, FMoW~\cite{christie2018functional}, and four datasets from the BREEDS benchmark~\cite{santurkar2020breeds}: Living-17, Nonliving-26, Entity-13, and Entity-30. Tiny-ImageNet is a compact subset of ImageNet comprising 200 classes. BREEDS (Benchmark for Robustness under Evolving Distribution Shifts) constructs subpopulation shifts by partitioning ImageNet classes hierarchically, where training and test sets contain different fine-grained subclasses within the same superclass, (\eg different dog breeds), simulating realistic deployment scenarios. For each source dataset, we evaluate performance on corresponding shifted test sets under distribution shift. 

For MNIST, we consider QMNIST~\cite{Yadav2019ColdCT}, USPS~\cite{Hull1994ADF}, and SVHN~\cite{Netzer2011ReadingDI} as shifted variants. For CIFAR-10, we use CIFAR-10v2~\cite{lu2020harder} as a natural shift (C10-N) and CIFAR-10-C~\cite{Hendrycks2016ABF} for synthetic corruptions (C10-S). For CIFAR-100, we use the standard test set as the natural shift benchmark (C100-N) and CIFAR-100-C~\cite{Hendrycks2016ABF} for synthetic corruptions (C100-S). For ImageNet, we assess natural shifts (IN-N) by ImageNet-V2~\cite{Recht2018DoCC} and ImageNet-Sketch~\cite{Wang2019LearningRG}, and synthetic corruption (IN-S) by ImageNet-C~\cite{Hendrycks2016ABF}. For Tiny-ImageNet, the natural shifts (IN200-N) include ImageNet-Reality~\cite{Hendrycks2020TheMF} and the corresponding 200 matching classes from ImageNet-V2 and ImageNet-Sketch. The synthetic shift (IN200-S) utilizes the 200 matching classes extracted from ImageNet-C. Note that the corruption benchmarks (CIFAR-10/100-C, ImageNet-C, ImageNet200-C) each contain 19 corruption types (\eg, Gaussian noise, motion blur, frost) with 5 severity levels per type, resulting in 95 distinct test conditions per dataset. FMoW naturally contains temporal and geographical distribution shifts and the out-of-distribution test data contain images from time periods and geographic regions unseen during training. For BREEDS datasets, we employ identical natural and synthetic shift protocols as ImageNet (V2 and C variants), models are evaluated on test sets containing either the training-time fine-grained categories (BREEDS-S) or novel categories from the same coarse-grained class (BREEDS-N), which isolates the subpopulation shift effects.

\section{Baselines}
\label{app:baselines}

We provide detailed formulations and implementation descriptions of all baselines compared in our experiments.

\paragraph{Importance Re-weighting (IM)} The IM method estimates the target error as a re-weighted source error, where the weights are obtained as the ratio between the densities of target and source data across confidence bins. Following \cite{Chen2021MandolineME} this effectively corresponds to using a single slice in the classifier confidence space.

\paragraph{Average Confidence (AC)} THe AC baseline directly estimates the target error by computing the average of one minus the maximum softmax confidence over the unlabeled target samples.

\paragraph{Difference of Confidence (DoC)} It is also known as DOC-Feat, which models the error as the difference between the source and target confidence distributions\cite{Guillory2021PredictingWC}. The formulation is $\hat{\varepsilon}_{\mathrm{DoC}} = \mathbb{E}_{x \sim D_S} \big[ \mathbb{I} \left[ \arg\max_{j \in \mathcal{Y}} f_{\theta}(j \mid x) \neq y \right] \big] + \mathbb{E}_{x \sim D_T} \big[ 1 - \max_{j \in \mathcal{Y}} f_{\theta}(y \mid x) \big] - \mathbb{E}_{x \sim D_S} \big[ 1 - \max_{j\in \mathcal{Y}} f_{\theta}(j \mid x) \big] $

\paragraph{Generalized Disagreement Equality (GDE)} It estimates the target error as the disagreement ratio
between the predictions of two independently trained models $f_{\theta}(x)$ and $f_{\theta'}(x)$
 on the target data\cite{Jiang2021AssessingGO}, which can be formulated as $\hat{\varepsilon}_{\mathrm{GDE}} = \mathbb{E}_{x \sim D_T} \Big[\mathbb{I}\!\left(\arg\max_{j \in \mathcal{Y}} f_{\theta}(j \!\mid\! x) \neq \arg\max_{j\in \mathcal{Y}} f_{\theta'}(j \!\mid\! x)\right)\Big].$

\paragraph{Average Thresholded Confidence (ATC)}
ATC estimates target error by identifying a threshold $t$ such that the fraction of source data points with scores below $t$
matches the validation error on source data. The target error is then estimated as the proportion of target examples falling below this threshold: $\hat{\varepsilon}_{\mathrm{ATC}}= \mathbb{E}_{x \sim D_T}\Big[\mathbb{I}\!\left( s(f_{\theta}(x)) < t \right)\Big]$, where $s(\cdot)$ denotes a scalar score function relating positively with the performance of the model. As proposed in~\cite{Garg2022LeveragingUD}, there are two variants considered: (1) \textbf{ATC-MC}, which uses the maximum softmax confidence $s_{\mathrm{MC}} = \max_{j\in \mathcal{Y}} f_{\theta}(j \!\mid\! x)$, and (2) \textbf{ATC-NE}, which uses the negative entropy score $s_{\mathrm{NE}} = -\sum_{j \in \mathcal{Y}} f_{\theta}(j \!\mid\! x) \log f_{\theta}(j\!\mid\! x)$.

\paragraph{Projection Normalization (ProjNorm)}
ProjNorm~\cite{Yu2022PredictingOE} originally proposed a parameter-space metric that quantifies the distributional shift between source and target domains. The original method does not directly estimate the target accuracy but instead demonstrates that the projection norm strongly correlates with the true target error. In our evaluation, since our goal is to directly predict model performance, we follow a practical implementation adapted from \cite{Lu2023CharacterizingOE}, which converts the original ProjNorm metric into an approximate accuracy estimator by comparing the output distributions between source and target data. 
This enables a fair, quantitative comparison under the same Mean Absolute Estimation Error (MAE) metric.

\paragraph{Confidence Optimal Transport (COT)} In \cite{Lu2023CharacterizingOE}, COT introduces an optimal-transport-based (OT) estimator that measures the Wasserstein distance between the empirical distribution of model confidence vectors on the unlabeled target set and the empirical source label distribution.
Unlike confidence-based estimators (\eg, Average Confidence) which may underestimate error by selecting the pseudo-label distribution as reference, COT uses the source label distribution under the assumption that $P_T(\vec{y}) \approx P_S(\vec{y})$. The paper further proposes \textbf{COTT} (COT with Thresholding), which learns a threshold on validation data and estimates error as the fraction of target samples whose per-sample transport costs exceed this threshold. 

\section{CWF on Additional Datasets}
\label{app:AdditionalCWF}

\begin{figure*}[t!]
    \centering
    \begin{subfigure}[b]{\columnwidth}
        \centering
        \includegraphics[width=\textwidth]{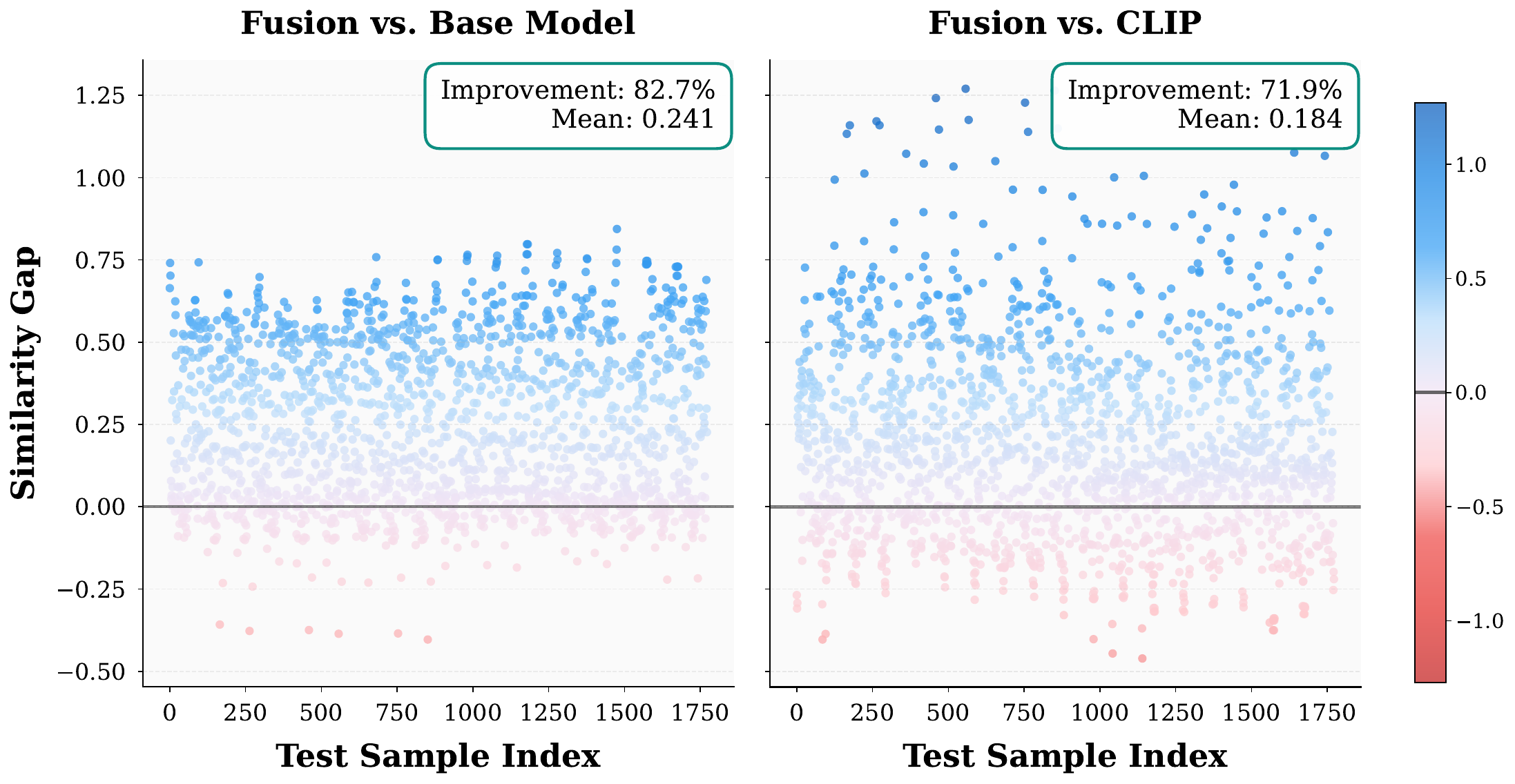}
        \caption{Same shift}
    \end{subfigure}
    \hfill
    \begin{subfigure}[b]{\columnwidth}
        \centering
        \includegraphics[width=\textwidth]{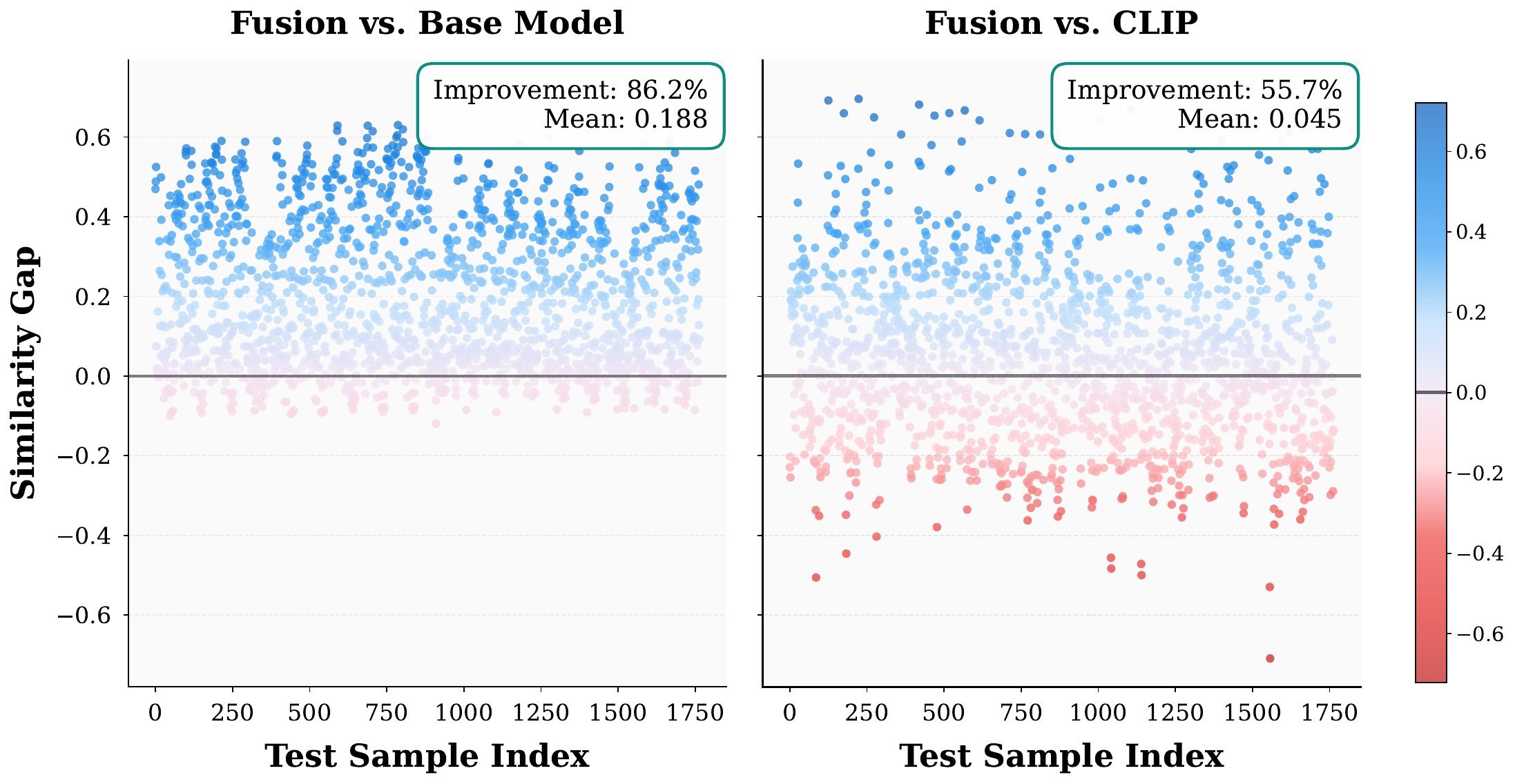}
        \caption{Novel shift}
    \end{subfigure}
    \caption{\textbf{Per-experiment \SAS (SAS) differences on Living-17}. We compare the confidence-weighted fusion against the base model (left column) and CLIP (right column) under (a) natural shifts and (b) synthetic corruptions. The y-axis shows $\Delta\text{SAS} = \text{SAS}_{\text{fusion}} - \text{SAS}_{\text{baseline}}$. Positive values indicate improved semantic alignment ($\Delta\text{SAS} > 0$), while negative values indicate degradation. The fusion largely enhances alignment under both conditions.}
    \label{fig:living17}
\end{figure*}
\begin{figure*}[t!]
    \centering
    \begin{subfigure}[b]{\columnwidth}
        \centering
        \includegraphics[width=\textwidth]{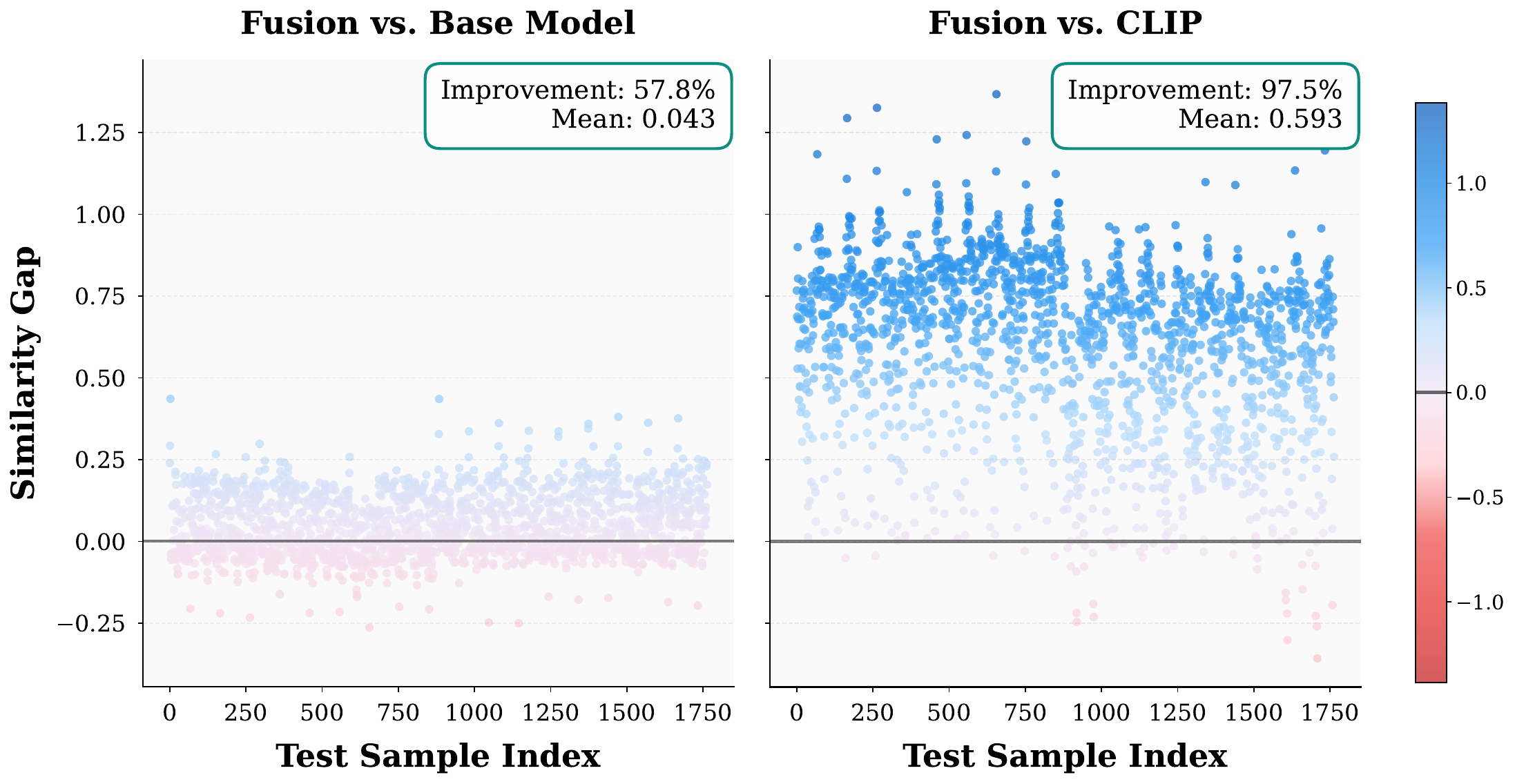}
        \caption{Same shift}
    \end{subfigure}
    \hfill
    \begin{subfigure}[b]{\columnwidth}
        \centering
        \includegraphics[width=\textwidth]{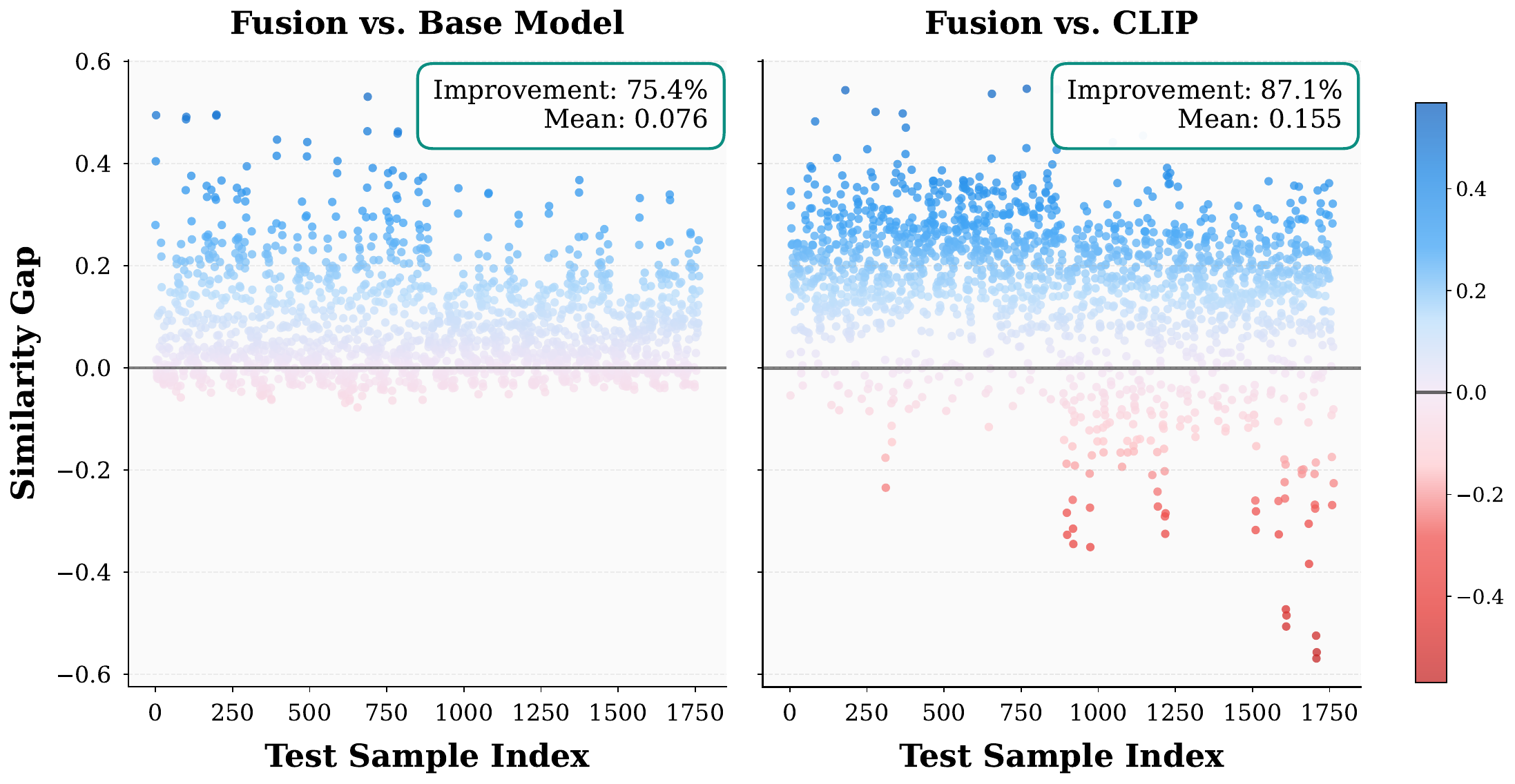}
        \caption{Novel shift}
    \end{subfigure}
    \caption{\textbf{Per-experiment \SAS (SAS) differences on Nonliving-26.} (a) same shifts and (b) novel shifts is the same with the Living-17. The fusion consistently achieves better semantic alignment than both the base model and CLIP.}
    \label{fig:nonliving26}
\end{figure*}
\begin{figure*}[t!]
    \centering
    \begin{subfigure}[b]{\columnwidth}
        \centering
        \includegraphics[width=\textwidth]{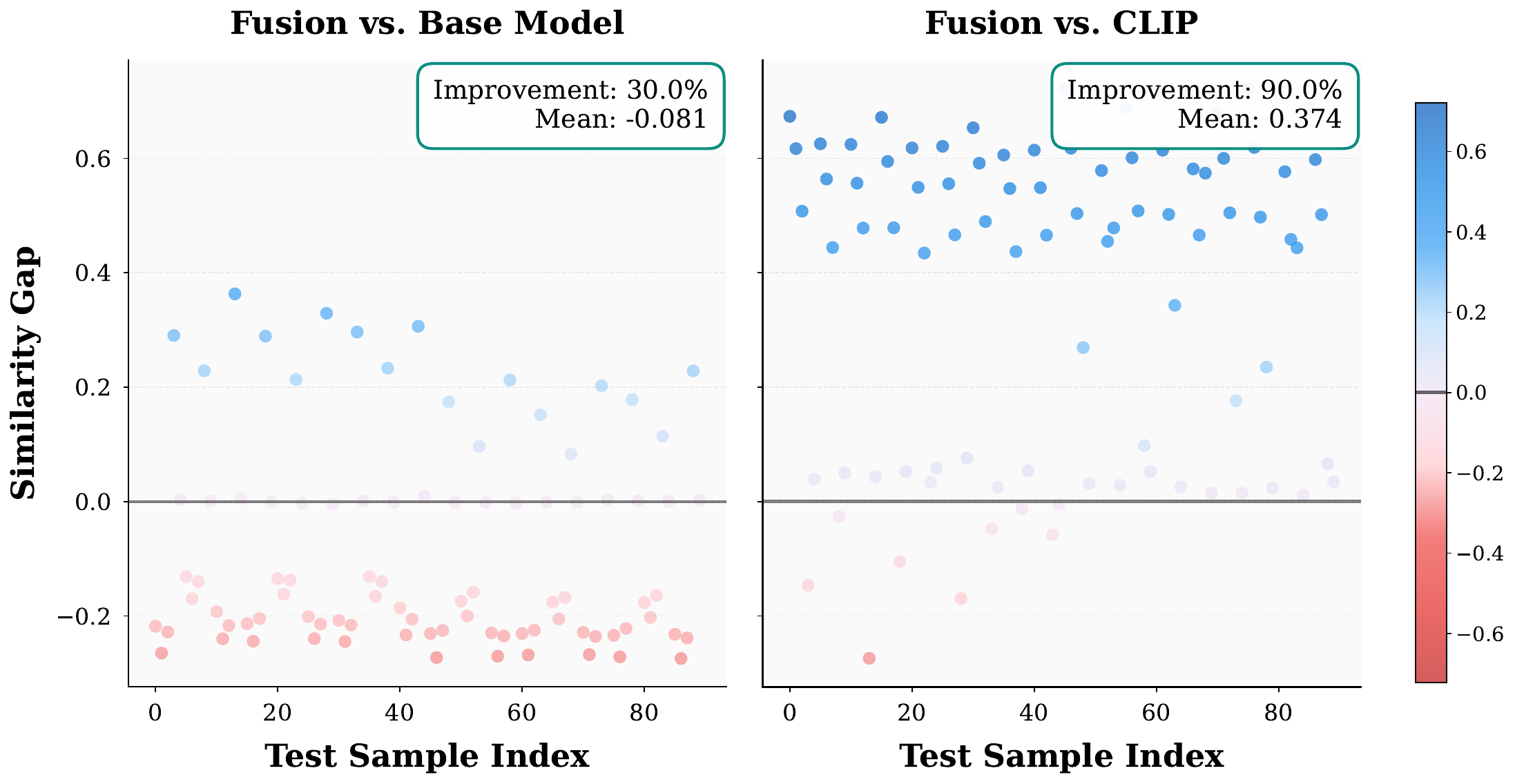}
    \end{subfigure}
    \hfill
    \begin{subfigure}[b]{\columnwidth}
        \centering
        \includegraphics[width=\textwidth]{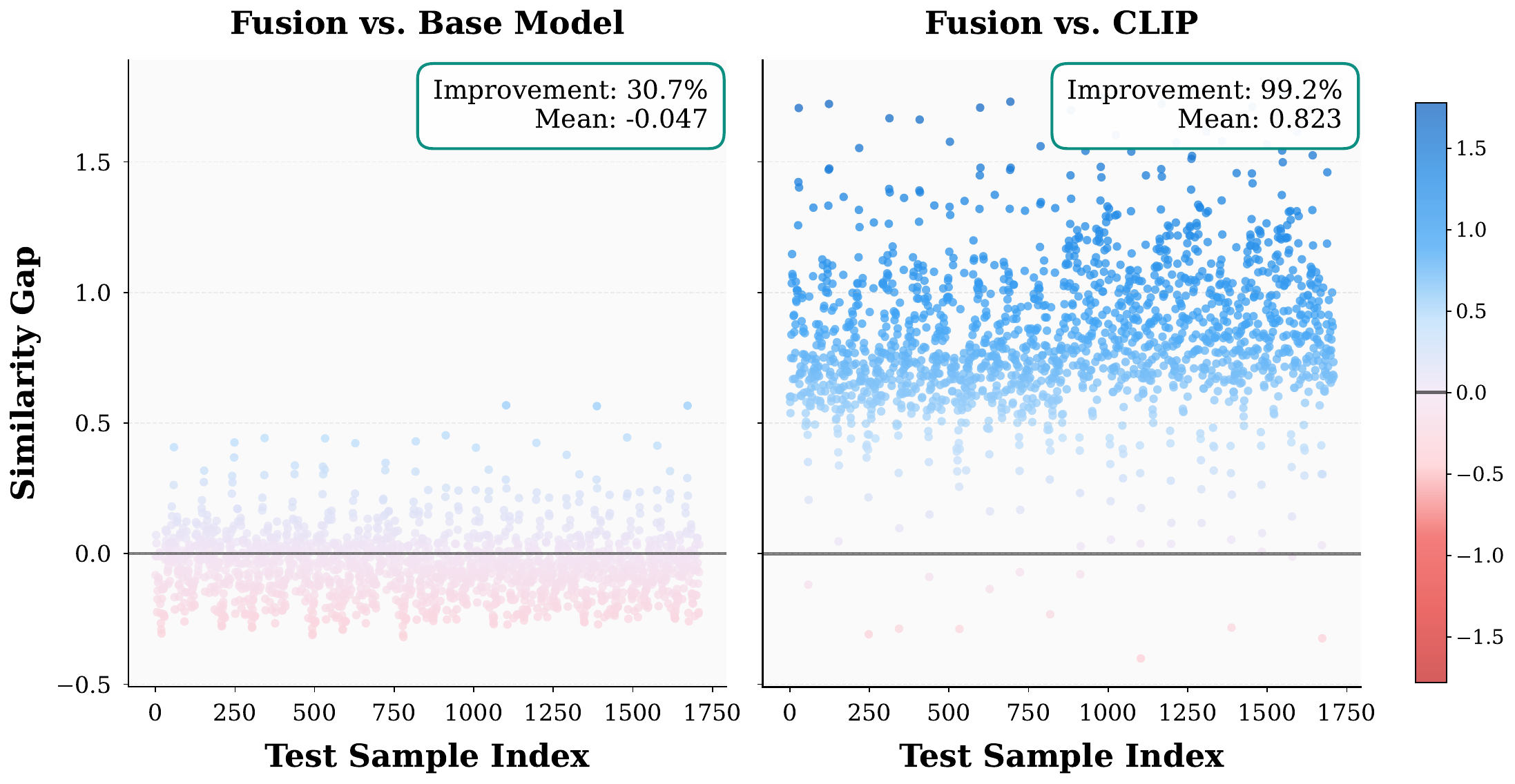}
    \end{subfigure}
    \caption{\textbf{Per-experiment \SAS (SAS) differences on Tiny-ImageNet.} (a) natural shifts and (b) synthetic corruptions is similar to the ImageNet. While the fused distribution on this dataset provides limited gains over the base model.}
    \label{fig:TinyImageNet}
\end{figure*}

To further validate the effectiveness of the confidence-weighted fusion (CWF), we conduct supplementary experiments on additional datasets including Tiny-ImageNet, Living-17, and Nonliving-26. We adopt the same evaluation protocol as described in $\text{Sec.}\,5.3$ to assess how well the fused predictions align semantically with the ground truth using \SAS (SAS). Note that this hierarchy-based evaluation is only applicable to datasets whose class labels conform to the WordNet taxonomy, which restricts our evaluation to the aforementioned datasets that satisfy this prerequisite. As shown in \cref{fig:living17,fig:nonliving26,fig:TinyImageNet}, the fused predictions on Living-17 and Nonliving-26 consistently exhibit markedly higher semantic consistency with the true labels than those of the base model. However, on Tiny-ImageNet, while the fused distribution substantially surpasses predictions of CLIP, it exhibits slightly degraded performance relative to the base model. 

This phenomenon suggests that poor performance of CLIP on this particular dataset adversely affects the fusion outcome, where the predictions of base model are compromised by weaker contributions of CLIP. This observation highlights a critical caveat: while CLIP exhibits remarkable zero-shot generalization capabilities, its knowledge coverage remains bounded. In certain specialized domains, its predictions can fall considerably short of those from task-specific base models. Nevertheless, this finding underscores the merit of our fusion-based approach over directly adopting CLIP predictions as an approximation of the ideal distribution. The integration of CLIP represents a calculated trade-off: while it may introduce adverse effects in limited scenarios, the confidence-weighted fusion mechanism predominantly yields improved predictive distributions across the majority of settings by effectively leveraging the complementary strengths of both models.

\section{Thresholding Strategy Ablation}
\label{app:ablation_thresholding}

\begin{table}[t]
\caption{\textbf{Ablation of thresholding strategy.} The performance difference between the standard FRAP and FRAP w/o Thresholding denoted as FRAP $_\text{(w/o thd)}$}
\label{tab:thresholdingResult}
\begin{center}
\small
\setlength{\tabcolsep}{6pt}
\begin{tabular}{llcc}
\toprule
\textbf{Dataset Family} & \textbf{Shift Type} & \textbf{FRAP} & \textbf{FRAP$_\text{(w/o thd)}$} \\
\midrule
MNIST & -- & \textbf{12.42}{\scriptsize$\pm$0.10} & 16.78{\scriptsize$\pm$0.04}\\
\midrule
\multirow{2}{*}{CIFAR-10} & C10-N & \textbf{2.14}{\scriptsize$\pm$0.02} & 6.13{\scriptsize$\pm$0.04}\\
& C10-S & \textbf{3.24}{\scriptsize$\pm$0.05} & 7.26{\scriptsize$\pm$0.05}\\
\midrule
\multirow{2}{*}{CIFAR-100} & C100-N & \textbf{4.00}{\scriptsize$\pm$0.03} & 7.97{\scriptsize$\pm$0.05}\\
& C100-S & 11.42{\scriptsize$\pm$0.10} & \textbf{8.53}{\scriptsize$\pm$0.06}\\
\midrule
\multirow{2}{*}{ImageNet} & IN-N & \textbf{2.16}{\scriptsize$\pm$0.03} & 12.31{\scriptsize$\pm$0.07}\\
& IN-S & \textbf{4.76}{\scriptsize$\pm$0.01} & 9.16{\scriptsize$\pm$0.06}\\
\midrule
\multirow{2}{*}{Tiny-ImageNet} & IN200-N & \textbf{7.46}{\scriptsize$\pm$0.04} & 14.7{\scriptsize$\pm$0.05}\\
& IN200-S & \textbf{9.98}{\scriptsize$\pm$0.04} & 12.37{\scriptsize$\pm$0.06}\\
\midrule
FMoW & -- & \textbf{3.42}{\scriptsize$\pm$0.02} & 15.74{\scriptsize$\pm$0.05}\\
\midrule
\multirow{4}{*}{BREEDS-S} & L17-S & \textbf{8.55}{\scriptsize$\pm$0.05} & 10.98{\scriptsize$\pm$0.06}\\
& NL26-S & \textbf{9.30}{\scriptsize$\pm$0.07} & 12.63{\scriptsize$\pm$0.07}\\
& E13-S & \textbf{8.29}{\scriptsize$\pm$0.05} & 11.69{\scriptsize$\pm$0.06}\\
& E30-S & \textbf{6.94}{\scriptsize$\pm$0.06} & 11.73{\scriptsize$\pm$0.07}\\
\midrule
\multirow{4}{*}{BREEDS-N} & L17-N & 5.37{\scriptsize$\pm$0.05} & \textbf{4.09}{\scriptsize$\pm$0.04}\\
& NL26-N & 6.66{\scriptsize$\pm$0.09} & \textbf{5.63}{\scriptsize$\pm$0.06}\\
& E13-N & \textbf{5.90}{\scriptsize$\pm$0.06} & 6.66{\scriptsize$\pm$0.05}\\
& E30-N & \textbf{5.54}{\scriptsize$\pm$0.07} & 6.4{\scriptsize$\pm$0.05}\\
\midrule
\rowcolor{gray!20}
\multicolumn{2}{l}{\textbf{Average}} & \textbf{6.53} & 10.04\\
\bottomrule
\end{tabular}
\end{center}
\label{apptb:thresholding}
\vspace{-0.5em}
\end{table}

As detailed in $\text{Sec.}\,5.4$, the thresholding strategy is introduced to address the inherent error of soft accuracy and the gap between the fused predictions and the true labels. This strategy transforms the continuous estimation problem into a more robust binary decision. By calibrating the threshold $\delta$ on the source validation set $\mathcal{D}_s$, we effectively align the fraction of samples predicted as incorrect (\ie, $\mathrm{Est}(x) < \delta$) with the observed error rate of the base model on $\mathcal{D}_s$. This process stabilizes the estimation by correcting the scale and offset between the raw prediction score and the actual error magnitude, relying on the relative ranking of scores rather than their absolute magnitudes.

To quantify the effect of the thresholding strategy, we perform an ablation study, which serves the formulation result in \cref{appeq:root} directly as the performance estimation, where the label distribution is replaced by the fused prediction produced through test-time calibration (TTC) and confidence-weighted fusion (CWF). This ablation allows us to isolate the performance gain specifically attributable to the thresholding step. 

As illustrated in \cref{apptb:thresholding}, ablation of the thresholding strategy results in a marked deterioration in estimation performance. Without thresholding, 15 out of 18 benchmark settings exhibit a higher estimation error, and the average MAE rises substantially from 6.53 to 10.04. This pronounced drop underscores the effect of thresholding in our framework. 

The thresholding strategy effectively alleviates the accumulated error arising from the soft-to-hard score difference and the inherent limitation of the fused predictions.

\begin{table}[t]
\caption{\textbf{Comparison with extensive baselines.} This table extends \cref{tab:comparison} with full results across all datasets. Background colors denote reference sources: {\setlength{\fboxsep}{1.5pt}\colorbox{pink!15}{CLIP}}, {\setlength{\fboxsep}{1.5pt}\colorbox{yellow!15}{SigLIP}}, {\setlength{\fboxsep}{1.5pt}\colorbox{green!10}{Random}}. Columns represent: (1) \textbf{Base}: reference pseudo-labels; (2) \textbf{Fix}: FRAP with fixed reference scaling; (3) \textbf{Dyna}: our full framework with Test-Time Calibration (TTC). \textbf{Random} reports performance using a pathological reference via Dirichlet distribution.}
\vspace{-3mm}
\label{tab:all_extensive_comparison}
\begin{center}
\setlength{\tabcolsep}{1.5pt}
\resizebox{\linewidth}{!}{
\begin{tabular}{>{\columncolor{gray!15}}l*{3}{>{\columncolor{pink!15}}c}*{3}{>{\columncolor{yellow!10}}c}>{\columncolor{green!10}}c} 
\toprule
\textbf{Data} & \textbf{Base} & \textbf{Fix} & \textbf{Dyna} & \textbf{Base} & \textbf{Fix} & \textbf{Dyna} & \textbf{Random} \\
\midrule
MNIST & 24.98$_{.21}$ & \textbf{11.13}$_{.10}$ & 12.42$_{.10}$ & 32.09$_{.26}$ & 13.54$_{.10}$ & \textbf{10.56}$_{.10}$ & 11.30$_{.10}$ \\
C10-N & 5.00$_{.15}$ & 8.48$_{.03}$ & \textbf{2.14}$_{.02}$ & \textbf{2.01}$_{.00}$ & 6.29$_{.02}$ & 7.54$_{.03}$ & 11.05$_{.04}$ \\
C10-S & 10.49$_{.13}$ & 8.01$_{.05}$ & \textbf{3.24}$_{.05}$ & 9.14$_{.09}$ & \textbf{5.67}$_{.04}$ & 7.68$_{.06}$ & 12.61$_{.07}$ \\
C100-N & 9.07$_{.08}$ & 10.90$_{.05}$ & \textbf{4.00}$_{.03}$ & \textbf{4.35}$_{.01}$ & 4.65$_{.05}$ & 10.22$_{.04}$ & 12.71$_{.06}$ \\
C100-S & \textbf{9.51}$_{.16}$ & 17.07$_{.10}$ & 11.42$_{.10}$ & \textbf{7.05}$_{.05}$ & 12.51$_{.09}$ & 17.87$_{.10}$ & 19.28$_{.10}$ \\
IN-S & 7.08$_{.29}$ & \textbf{3.11}$_{.02}$ & 4.76$_{.01}$ & 4.02$_{.02}$ & \textbf{3.57}$_{.02}$ & 3.74$_{.02}$ & 2.99$_{.02}$ \\
IN-N & 9.25$_{.24}$ & \textbf{1.43}$_{.01}$ & 2.16$_{.03}$ & 3.80$_{.02}$ & \textbf{0.88}$_{.01}$ & 1.15$_{.01}$ & 1.97$_{.02}$ \\
IN200-S & 12.97$_{.23}$ & \textbf{4.95}$_{.04}$ & 9.98$_{.04}$ & 8.74$_{.02}$ & 6.33$_{.04}$ & \textbf{6.14}$_{.05}$ & 4.86$_{.04}$ \\
IN200-N & 11.97$_{.19}$ & \textbf{5.32}$_{.06}$ & 7.46$_{.04}$ & 6.53$_{.04}$ & \textbf{4.31}$_{.05}$ & 5.43$_{.07}$ & 5.91$_{.07}$ \\
FMoW & 40.86$_{.10}$ & \textbf{3.39}$_{.03}$ & 3.42$_{.02}$ & 41.27$_{.07}$ & \textbf{2.20}$_{.02}$ & 2.70$_{.03}$ & 2.48$_{.03}$ \\
L17-S & \textbf{4.74}$_{.23}$ & 6.15$_{.05}$ & 8.55$_{.05}$ & \textbf{4.66}$_{.04}$ & 6.33$_{.05}$ & 6.81$_{.06}$ & 5.80$_{.06}$ \\
NL26-S & 7.79$_{.26}$ & \textbf{5.64}$_{.05}$ & 9.30$_{.07}$ & 7.16$_{.04}$ & \textbf{5.28}$_{.04}$ & 7.60$_{.09}$ & 7.24$_{.08}$ \\
E13-S & 10.39$_{.19}$ & \textbf{4.93}$_{.05}$ & 8.29$_{.05}$ & 12.38$_{.05}$ & \textbf{4.55}$_{.04}$ & 5.78$_{.07}$ & 5.75$_{.06}$ \\
E30-S & 10.53$_{.26}$ & \textbf{4.07}$_{.05}$ & 6.94$_{.06}$ & 10.02$_{.05}$ & \textbf{3.92}$_{.05}$ & 5.81$_{.08}$ & 5.26$_{.07}$ \\
L17-N & 1.82$_{.22}$ & \textbf{5.70}$_{.05}$ & 5.37$_{.05}$ & \textbf{1.65}$_{.02}$ & 5.26$_{.04}$ & 6.74$_{.07}$ & 10.20$_{.07}$ \\
NL26-N & \textbf{0.85}$_{.18}$ & 5.74$_{.05}$ & 6.66$_{.09}$ & \textbf{1.02}$_{.01}$ & 4.91$_{.04}$ & 8.48$_{.10}$ & 9.04$_{.09}$ \\
E13-N & \textbf{5.19}$_{.21}$ & 6.98$_{.06}$ & 5.90$_{.06}$ & \textbf{6.14}$_{.03}$ & 6.43$_{.05}$ & 8.50$_{.07}$ & 8.70$_{.07}$ \\
E30-N & \textbf{3.19}$_{.22}$ & 6.01$_{.05}$ & 5.54$_{.07}$ & \textbf{3.77}$_{.02}$ & 5.08$_{.05}$ & 9.05$_{.09}$ & 8.70$_{.07}$ \\
\midrule
\textbf{Avg} & 10.32 & 6.61 & \textbf{6.53} & 9.21 & \textbf{5.65} & 7.32 & 8.10 \\
\bottomrule
\end{tabular}
}
\end{center}
\vspace{-1em} 
\end{table}


\end{document}